\documentclass{article}
\PassOptionsToPackage{numbers,compress}{natbib}
\usepackage[preprint]{neurips_2026} 
\usepackage{placeins}
\usepackage{enumitem}
\usepackage{multirow}

\usepackage{amsmath,amsfonts,bm}

\def\eqref#1{equation~\ref{#1}}

\def\1{\bm{1}}

\DeclareMathAlphabet{\mathsfit}{\encodingdefault}{\sfdefault}{m}{sl}
\SetMathAlphabet{\mathsfit}{bold}{\encodingdefault}{\sfdefault}{bx}{n}

\usepackage{fp}
\usepackage[utf8]{inputenc}
\usepackage[T1]{fontenc}
\usepackage{hyperref}
\hypersetup{colorlinks=true}
\usepackage{url}
\usepackage{booktabs}
\usepackage{amsfonts}
\usepackage{nicefrac}
\usepackage{microtype}
\usepackage{xcolor}
\usepackage{subcaption}
\usepackage{adjustbox} 
\usepackage{calc}
\usepackage{hyperref}
\usepackage{url}
\usepackage{booktabs}
\usepackage{amsfonts}
\usepackage{nicefrac}
\usepackage{microtype}
\usepackage{wrapfig}
\usepackage{graphicx}
\usepackage{caption}
\usepackage{subcaption}
\usepackage{xcolor}
\usepackage{xspace}
\usepackage{amsmath}
\usepackage{amssymb}
\usepackage{mathtools}
\usepackage{amsthm}
\usepackage{tabularx}
\usepackage[most,skins,theorems]{tcolorbox}
\usepackage{tikz}
\usepackage{framed}
\usepackage{pifont}
\tcbset{
  aibox/.style={
    width=\linewidth,
    top=8pt,
    bottom=4pt,
    colback=blue!6!white,
    colframe=black,
    colbacktitle=black,
    enhanced,
    center,
    attach boxed title to top left={yshift=-0.1in,xshift=0.15in},
    boxed title style={boxrule=0pt,colframe=white,},
  }
}
\newtcolorbox{AIbox}[2][]{aibox,title=#2,#1}
\usepackage{fontawesome5}  
\usepackage{booktabs}
\usepackage{multirow}
\usepackage{tikz}
\usepackage{pgfplots}
\pgfplotsset{compat=1.18}
 \usepackage{wrapfig}
\usepackage{graphicx}
\usepackage{siunitx}
\usepackage{enumitem}
\usepackage{multirow}

\usepackage{etoolbox}
\usepackage{fp}
\usepackage{amsmath}

\usepackage{amssymb}
\usepackage{pifont}

\usepackage{gensymb}
\usepackage{calc}
\usepackage{booktabs}
\usepackage{amsfonts}

\usepackage{graphicx}
\usepackage{caption}
\usepackage{subcaption}
\usepackage{booktabs}
\usepackage{multirow}
\usepackage{colortbl}
\usepackage{float}
\usepackage[super]{nth}
\usepackage{tikz}
\usepackage{pgfplots}
\usetikzlibrary{shadows.blur,shapes.symbols,calc,decorations.pathreplacing}
\usepackage{graphicx}
\usepackage{siunitx}
\usepackage{enumitem}
\usepackage{multirow}
\usepackage{fp}
\usepackage{amsmath}
\usepackage{amssymb}
\usepackage{xcolor}
\usepackage{algorithm}
\usepackage{algorithmicx}
\usepackage{algpseudocode}
\usepackage{wrapfig}
\usepackage{pifont}
\usepackage{booktabs} 
\usepackage{caption}  
\usepackage{makecell}
\usepackage{fontawesome5}

\newcommand{\Gray}[0]{\rowcolor{gray!20}}
\newcommand{\Lgray}[0]{\rowcolor{gray!10}}

\newcommand{\icohalf}{\textcolor{darkorange}{\ding{51}\kern-0.65em\ding{55}}}
\usepackage{tikz}
\usepackage{pifont}

\usepackage{pifont}
\usepackage{bm}
\usepackage[normalem]{ulem}
\usepackage{afterpage}
\usepackage{float}
\usepackage{fontawesome5} 
\definecolor{indianred}{rgb}{0.8, 0.36, 0.36}
\definecolor{bleudefrance}{rgb}{0.19, 0.55, 0.91}
\definecolor{forestgreen}{rgb}{0.0, 0.5, 0.0}
\definecolor{ashgrey}{rgb}{0.7, 0.75, 0.71}
\definecolor{darkorange}{rgb}{1.0, 0.55, 0.0}
\definecolor{darkorchid}{rgb}{0.6, 0.2, 0.8}

\definecolor{BurntOrange}{rgb}{0.8, 0.33, 0.0}
\usepackage{xcolor}

\definecolor{mycolor_green}{HTML}{E6F8E0}
\definecolor{backred}{RGB}{255, 190, 190}
\definecolor{red}{RGB}{139, 0, 0}
\definecolor{purple}{HTML}{E6F8E0}
\usepackage{framed}

\definecolor{verylightgray}{HTML}{E6F8E0}

\definecolor{lightgray}{gray}{0.95}

\usepackage{multirow}
\usepackage{colortbl}
\usepackage{makecell}
\usepackage{caption}
\usepackage{adjustbox}
\makeatletter
  \newcommand\figcaption{\def\@captype{figure}\caption}
  \newcommand\tabcaption{\def\@captype{table}\caption}
\makeatother
\usepackage[most]{tcolorbox}

\usepackage{tabularx}

\usepackage{xcolor}
\usepackage{hyperref}
\hypersetup{colorlinks=true}
\definecolor{mycolor}{RGB}{128, 0, 255}
\definecolor{wfx}{RGB}{128, 0, 255}
\definecolor{wdcolor}{RGB}{128, 0, 255}

\definecolor{wdqcolor}{RGB}{255, 0, 0}

\definecolor{wdqcolor}{RGB}{255, 0, 0}

\title{Beyond Zooming: Learning Multi-Tool Visual Reasoning for Ultra-High-Resolution Remote Sensing}

\author{
\parbox{\dimexpr\textwidth-14pt\relax}{\centering
\textbf{Fengxiang Wang\textsuperscript{1,*}, Jiangnan Huang\textsuperscript{2,*}, Mingshuo Chen\textsuperscript{1}, Yueying Li\textsuperscript{1},} \\
\textbf{Yang Shi\textsuperscript{1}, Junwei Luo\textsuperscript{2}, Haoyu Wang\textsuperscript{3,\textdagger}, Yansheng Li\textsuperscript{2}, Jing Zhang\textsuperscript{2},} \\
\textbf{Haiyan Zhao\textsuperscript{3,\textdagger}, Wenjing Yang\textsuperscript{1}} \\
\textsuperscript{1} National University of Defense Technology \quad
\textsuperscript{2} Wuhan University \quad
\textsuperscript{3} Tsinghua University \\
\textsuperscript{*} Equal contribution \quad
\textsuperscript{\textdagger} Corresponding authors
}
}

\begin{document}
\emergencystretch=3em
\maketitle
\begin{abstract}
Ultra-high-resolution (UHR) remote-sensing (RS) imagery provides fine-grained Earth-observation evidence over city-scale scenes, but poses a fundamental challenge for multimodal large language models (MLLMs): task-relevant evidence is often sparse, local, and spatially dispersed across extremely large visual contexts. A natural solution is to equip MLLMs with zoom-in tools for active local inspection. However, through a pilot study on XLRS-Bench, we find that zoom-in is only partially effective: it resolves easy and medium-level tasks with locally recoverable evidence, but saturates on hard cases requiring global search, multi-region comparison, path planning, or dispersed-evidence reasoning. Motivated by this finding, we move beyond single-tool zoom-in and introduce \textbf{GeoMTVR}, a large-scale Geospatial Multi-Tool Visual Reasoning dataset built from wide-area satellite imagery. GeoMTVR contains 13K UHR VQA samples with interleaved reasoning trajectories, diverse visual tool calls, and returned visual observations, enabling models to learn question decomposition, tool selection, regional inspection, object-level grounding, auxiliary visual reasoning, and cross-tool evidence integration. Beyond supervised fine-tuning, we propose a tool-attention-focused reinforcement learning algorithm that concentrates optimization on critical tool-use decisions, including when to invoke tools, which tool to select, where to apply it, and how to interpret tool outputs. By combining SFT on GeoMTVR with our RL algorithm, we develop \textbf{GeoLens}, a multi-tool visual reasoning MLLM for UHR RS. Experiments show that GeoLens consistently outperforms direct reasoning and single-tool zoom-in baselines, achieving stronger accuracy, better evidence grounding, and more efficient tool-use trajectories. Our results suggest that UHR RS MLLMs should evolve from passive image understanding and single-operation exploration toward active, task-adaptive, multi-tool visual reasoning.
Datasets and code were released at \href{https://github.com/Fengxiang23/GeoLens}{\textcolor{red}{GeoLens}}.
\end{abstract}

\section{Introduction}
\label{section1}

\begin{figure*}[t]
\vspace{2mm}
\centering
\includegraphics[width=1.0\textwidth]{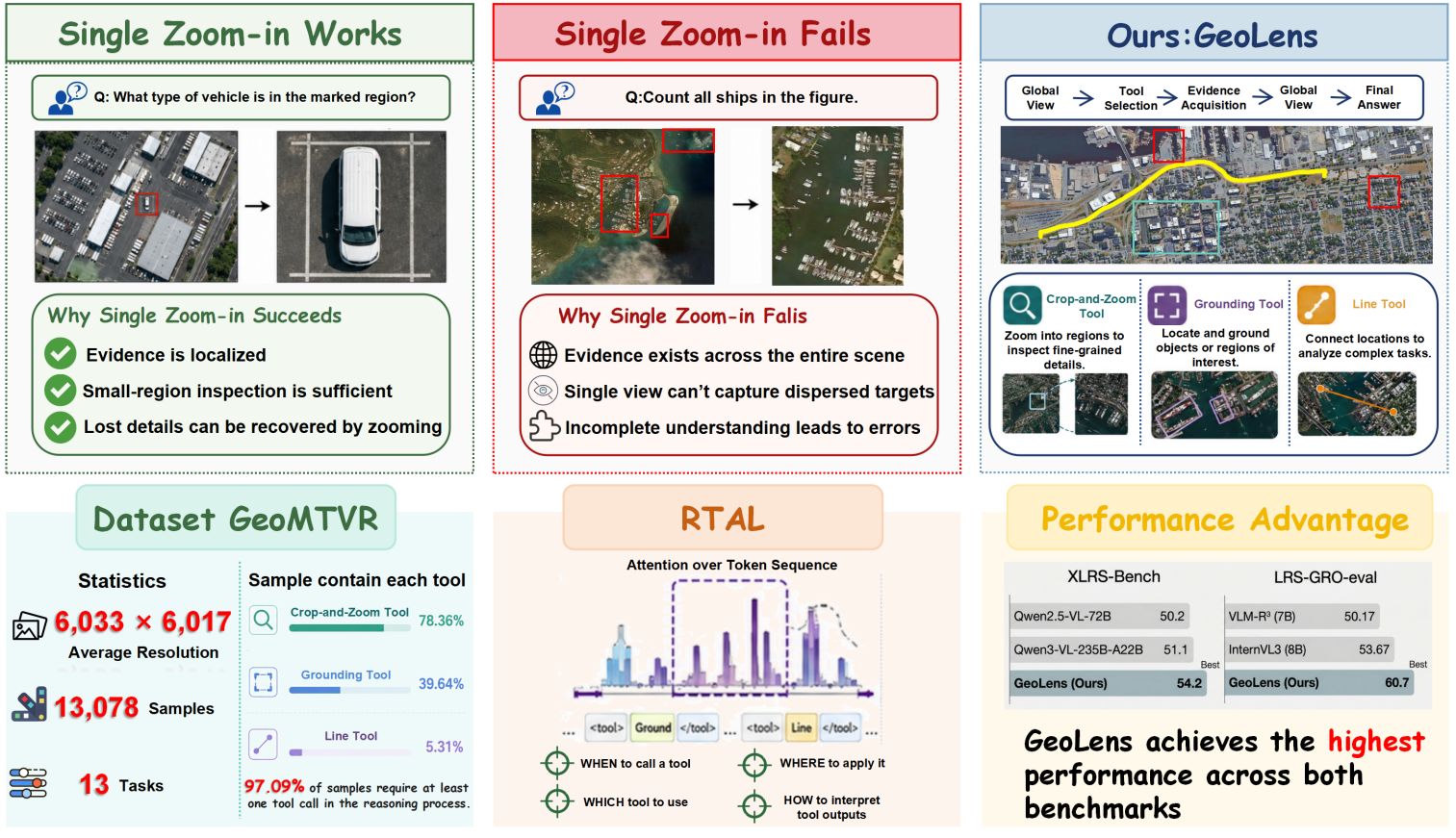}
\captionsetup{skip=3pt}
\caption{\textbf{Overview of GeoLens.} GeoLens enables multi-tool visual reasoning for ultra-high-resolution remote-sensing imagery.}
\label{fig:intro}
\vspace{-8mm}
\end{figure*}

Ultra-high-resolution (UHR) remote-sensing (RS) imagery greatly increases the observable detail of the Earth's surface. Unlike conventional natural images, a single UHR satellite image often covers city-scale or regional-scale areas while preserving small objects, local structures, and long-range geographic relations. This makes UHR RS imagery an important yet challenging domain for multimodal large language models (MLLMs)~\citep{wang2025geollava8kscalingremotesensingmultimodal,xlrs-bench,dang2025benchmark}.

Recent advances in MLLMs have substantially improved visual understanding and reasoning in both general and remote-sensing domains~\citep{geochat,earthgpt,rsgpt,zhan2025skyeyegpt}. However, UHR RS understanding remains far from solved. The core bottleneck is visual evidence acquisition: the model must identify task-relevant evidence from extremely large visual contexts, where useful regions are often sparse, local, and spatially dispersed. Directly compressing the entire image into a global representation may discard fine-grained details, while exhaustive local inspection is computationally inefficient and difficult to learn~\citep{lrsvqa,wang2025geollava8kscalingremotesensingmultimodal}.


A natural solution is to equip MLLMs with visual tools, especially crop-and-zoom, so that they can actively inspect high-resolution local regions during reasoning. Given an image and a question, the model interleaves textual reasoning with visual actions and decides whether to continue reasoning or invoke a zoom-in tool to acquire finer local evidence~\citep{zheng2025deepeyes,hong2025deepeyesv2,shen2025zoomeye}. This paradigm has recently attracted attention in both general-domain and RS MLLMs~\citep{wang2025vicot,liu2025zoomearth,wang2026geoeyes}. However, recent general-domain analyses suggest that zoom-in is not a universally effective operation: its benefit depends heavily on whether the task can be solved through localized evidence recovery, and indiscriminate zoom-in may introduce noisy or incomplete evidence~\citep{hou2025codev,wei2026zooming,ma2026does}. In contrast, recent UHR RS studies emphasize zoom-in as a key mechanism for resolving fine-grained details in large satellite images~\citep{liu2025zoomearth,wang2026geoeyes}. This discrepancy raises a central question:

\begin{tcolorbox}[top=1pt, bottom=1pt, left=1pt, right=1pt]
\textit{Is single-tool zoom-in sufficient for UHR remote-sensing reasoning, or do UHR RS MLLMs require more diverse visual tools for active evidence acquisition?}
\end{tcolorbox}

To answer this question, we first conduct a pilot study on XLRS-Bench using a zoom-in-enabled reinforcement learning framework~\citep{li2026ral}. Our analysis shows that zoom-in is useful, but its benefit is highly task-dependent. It mainly improves easy and medium-level tasks whose missing evidence is locally recoverable, such as object recognition, local counting, and attribute identification. However, it saturates on hard cases that require global search, multi-region comparison, long-range spatial reasoning, or heterogeneous perceptual operations, such as global counting, path planning, and dispersed-evidence reasoning. These results suggest that the limitation is not simply an insufficient number of zoom-in calls. Rather, single-tool zoom-in provides only one form of evidence acquisition, while UHR RS reasoning often requires models to decide which visual operation to perform, where to apply it, and how to integrate evidence returned by different tools.

Motivated by this finding, we move beyond single-tool zoom-in and study multi-tool visual reasoning for UHR RS. Specifically, we consider three complementary visual tools: \texttt{crop\_and\_zoomin} for progressive local inspection, \texttt{grounding} for precise object-level localization, and \texttt{line} for auxiliary-line drawing in spatial relation and path-oriented reasoning~\citep{hu2024visual,liu2023grounding,ma2026blueprints}. To support this setting, we construct \textbf{GeoMTVR} (\textbf{Geo}spatial \textbf{M}ulti-\textbf{T}ool \textbf{V}isual \textbf{R}easoning), a large-scale supervised fine-tuning dataset from wide-area satellite imagery. Each sample in GeoMTVR provides an interleaved trajectory of textual reasoning, visual tool calls, and returned visual observations. Different from existing high-resolution RS VQA datasets, which are mostly static question-answer datasets or zoom-in-only tool-interaction datasets~\citep{wang2025geollava8kscalingremotesensingmultimodal,lrsvqa,liu2025zoomearth,wang2026geoeyes}, GeoMTVR enables models to learn question decomposition, tool selection, regional inspection, object-level grounding, auxiliary visual reasoning, and cross-tool evidence integration.

However, supervised learning alone may imitate tool-use traces without fully optimizing tool-use decisions. In UHR RS reasoning, the key challenge is not only generating correct answers, but also allocating model attention to critical tool-related decisions, including when to call a tool, which tool to select, where to apply it, and how to interpret the returned observation. To this end, we propose Reinforced Tool Attention Learning (RTAL), a tool-attention-focused reinforcement learning algorithm inspired by attention-level post-training objectives~\citep{li2026ral}. RTAL places stronger optimization emphasis on tokens and attention patterns associated with tool-use decisions. By concentrating policy updates around tool invocation and tool-output interpretation, RTAL encourages more adaptive and efficient evidence acquisition, rather than uniformly reinforcing all generated tokens.

Based on GeoMTVR and RTAL, we develop \textbf{GeoLens}, a multi-tool visual reasoning MLLM for UHR remote sensing. GeoLens is first trained via supervised fine-tuning (SFT) on multi-tool reasoning trajectories and then further optimized with RTAL. As a result, the model learns to actively select visual tools, inspect relevant regions, and integrate evidence from heterogeneous observations. Experiments on UHR RS reasoning benchmarks show that GeoLens consistently outperforms direct reasoning and single-tool zoom-in baselines, achieving stronger accuracy, better evidence grounding, and more efficient tool-use trajectories.

In summary, our main contributions are as follows:
\begin{itemize}
    \item[(1)] We systematically analyze the effectiveness and limitations of single-tool zoom-in in UHR RS reasoning. Our pilot study shows that zoom-in mainly solves localized easy and medium-level tasks, such as object recognition, local counting, and attribute identification, but saturates on hard cases requiring global coverage, multi-region comparison, or structured spatial reasoning.
    
    \item[(2)] We construct \textbf{GeoMTVR}, a large-scale UHR RS multi-tool visual reasoning dataset with interleaved textual reasoning, three types of visual tool calls, and returned visual observations. To our knowledge, it is the first dataset to support multi-tool visual reasoning in UHR RS, covering crop-and-zoom, grounding, and auxiliary-line drawing.

    \item[(3)] We propose RTAL and develop \textbf{GeoLens}, a multi-tool visual reasoning MLLM for UHR remote sensing. By combining SFT on GeoMTVR with RTAL, GeoLens learns to actively select visual tools, inspect relevant regions, and integrate heterogeneous visual evidence, achieving consistent improvements over direct reasoning and single-tool zoom-in baselines.
\end{itemize}

\section{Related Work}
\label{section2}
\vspace{-1mm}
\textbf{Remote Sensing Large Multimodal Models.}
Recent advances in large multimodal models (MLLMs) have promoted remote sensing (RS) understanding across multi-task reasoning~\citep{rsgpt,zhan2025skyeyegpt}, captioning and retrieval~\citep{silva2024llmremote}, interactive dialogue~\citep{earthgpt}, unified instruction following~\citep{geochat,lhrsbot,li2024lhrs}, multi-sensor analysis~\citep{shu2025earthmind,soni2024earthdial,wang2026earthvl}, grounded understanding~\citep{wang2025ringmogpt,zhang2024earthmarker}, and mixture-of-experts modeling~\citep{LIU2026113717}. As RS imagery moves toward ultra-high-resolution (UHR) inputs, the prohibitive number of visual tokens becomes a central bottleneck. Existing solutions mainly follow two directions. One line reduces visual tokens through pruning or evidence selection, using adaptive clustering, attention-guided cropping, or reference-guided retrieval to remove redundant regions~\citep{wang2025geollava8kscalingremotesensingmultimodal,shalevarkushin2025imageragdynamicimageretrieval}. These methods improve efficiency but often depend on hand-crafted rules or static selection strategies, which may not generalize to diverse VQA tasks. Another line equips RS MLLMs with visual tools for iterative perception. Representative systems such as ZoomEarth~\citep{liu2025zoomearth}, VICoT-Agent~\citep{wang2025vicot}, and GeoEyes~\citep{wang2026geoeyes} progressively refine visual evidence, typically through zoom-in operations over regions of interest. However, current tool-based RS models still rely mainly on zoom-in as the dominant interaction form, leaving complex UHR scenarios underexplored, where evidence may be distributed across different scales, locations, and semantic contexts.

\textbf{Remote Sensing Vision-Language Datasets.}
The RS community has developed many vision-language datasets to support multimodal learning, including large-scale image-text resources for geo-foundation models~\citep{yuan2024chatearthnet}. Early datasets such as RSVQA~\citep{lobry2020rsvqa}, RS-IVQA~\citep{rsivqa}, and EarthVQA~\citep{wang2024earthvqa} construct VQA samples from geographic annotations or automatically generated questions, while RSSA~\citep{h2rsvlm}, SkySenseGPT~\citep{skysensegpt}, VHM~\citep{pang2025vhm}, and VRSBench~\citep{vrsbench} expand instruction-following and task coverage. However, these datasets are generally limited in spatial resolution. Recent UHR datasets, including SuperRS-VQA~\citep{wang2025geollava8kscalingremotesensingmultimodal}, HighRS-VQA~\citep{wang2025geollava8kscalingremotesensingmultimodal}, LRS-VQA~\citep{lrsvqa}, XLRS-Bench~\citep{xlrs-bench}, and RSHR-Bench~\citep{dang2025benchmark}, push RS evaluation toward the 8K scale. Tool-augmented datasets such as GeoEyes UHR-CoZ~\citep{wang2026geoeyes} and LRS-GRO~\citep{liu2025zoomearth} further introduce external visual operations into RS reasoning. Nevertheless, existing datasets still provide limited support for multi-step and multi-tool reasoning in UHR settings. Most UHR datasets focus on single-step perception or evaluation, while current tool-augmented datasets mainly rely on shallow zoom-in interactions. General-domain multi-tool instruction data is also not tailored to RS imagery or UHR inputs, creating a gap for training MLLMs to perform complex thinking with visual tools in real-world RS scenarios.

\textbf{General Thinking with Image.}
Chain-of-Thought reasoning has improved the reasoning ability of vision-language models, but text-only CoT still depends on single-pass image perception and can miss fine visual evidence. Recent ``thinking with image'' methods address this issue by allowing models to interact with images during reasoning~\citep{su2025thinking}. Visual Sketchpad~\citep{hu2024visual} introduces image manipulation tools such as auxiliary lines, while structured visual representations further improve spatial reasoning~\citep{ma2026blueprints}. Visual Tool Agent~\citep{huang2025visualtoolagent} and SpaceTools~\citep{chen2025spacetoolstoolaugmentedspatialreasoning} show that models can learn to select tools for multimodal and spatial reasoning. Other works use zoom-in or crop-based operations to improve small-object recognition and fine-grained perception~\citep{kumar2025reinforcing,carvalho2025cropvlm,zhang2025chain}. Recent agentic VLMs further combine tool invocation, visual action trajectories, and reinforcement learning to acquire visual evidence during reasoning~\citep{zheng2025deepeyes,hong2025deepeyesv2,shen2025zoomeye,SenseNova-MARS,pixelreasoner}. In RS, vision-interleaved reasoning has also begun to connect tool use with geospatial interpretation~\citep{wang2025vicot}. Meanwhile, several studies indicate that the effectiveness of zoom-in depends on training data, reward design, and tool-use policy rather than the operation itself~\citep{hou2025codev,wei2026zooming,ma2026does}. Existing methods, however, are mostly designed for natural images or single-tool settings. They do not fully address UHR RS imagery, where useful evidence is sparse, spatially dispersed, and often requires domain-aware multi-step visual reasoning.

\vspace{-3mm}
\section{Dataset}\label{section3}

Recent advances in remote sensing multimodal large models have led to substantial VQA datasets, including SuperRS-VQA~\citep{wang2025geollava8kscalingremotesensingmultimodal}, HighRS-VQA~\citep{wang2025geollava8kscalingremotesensingmultimodal}, RSVQA-HR~\citep{lobry2020rsvqa}, UHR-CoZ~\citep{wang2026geoeyes}, and LRS-GRO~\citep{liu2025zoomearth}. 
However, existing high-resolution RS VQA datasets mainly fall into two categories: static question--answer datasets and tool-interactive datasets centered on zoom-in operations.
They therefore provide limited support for diverse visual tool use, which is essential in ultra-high-resolution (UHR) scenarios where reasoning often requires coordinated inspection, localization, and spatial comparison across multiple regions.
Table~\ref{tab:dataset-compare} compares our dataset with existing datasets.

\begin{table}[H]
\centering
\vspace{-2mm}
\footnotesize
\caption{\textbf{Comparison with RS UHR VQA datasets.}}
\label{tab:dataset-compare}
\resizebox{0.86\textwidth}{!}{
\begin{tabular}{lccl}
\toprule
\textbf{Dataset} & \textbf{Avg. Resolution} & \textbf{Tool Types} & \textbf{VQA Volume} \\
\midrule
RSVQA~\citep{lobry2020rsvqa} & 512$\times$512 & None & 111,134 \\
RSVQA-HR~\citep{lobry2020rsvqa} & 1,024$\times$1,024 & None & 1,066,316 \\
LRS-GRO~\citep{liu2025zoomearth} & 5,000$\times$5,000 & Zoom-in & 13,245 \\
UHR-CoZ~\citep{wang2026geoeyes} & 2,178$\times$2,051 & Zoom-in & 25,467 \\
\midrule
\textbf{GeoMTVR} & \textbf{6,033$\times$6,017} & \textbf{\makecell[l]{Zoom-in,\\ Grounding, Line}} & \textbf{13,078} \\
\bottomrule
\end{tabular}}
\vspace{-4mm}
\end{table}

To fill this gap, we follow recent paradigms such as ViLaSR~\citep{wu2025reinforcingspatialreasoningvisionlanguage}, which introduce tool invocation and visual-process annotation, and extend them to UHR satellite imagery.
We design a staged, semi-automatic, process-driven annotation framework that progressively converts existing UHR RS VQA samples into interpretable tool-use reasoning trajectories.
Through multi-stage collaboration, the framework produces high-quality annotations involving diverse visual tools, including crop-and-zoom, grounding, and auxiliary-line operations.
Using this pipeline, we construct approximately 13K high-resolution VQA samples with an average resolution of about 9K $\times$ 9K, each enriched with detailed visual tool-use reasoning traces.

\begin{table}[H]
\centering
\footnotesize
\caption{\textbf{Pilot Experiments of Dataset.} We only modify the SFT data. Accuracy is reported based on results from XLRS-Bench~\citep{xlrs-bench}.}
\label{tab:dataset-motivation}
\resizebox{\textwidth}{!}{
\begin{tabular}{clllllc}\toprule
\multicolumn{1}{l}{\multirow{2}{*}{Model}} & \multicolumn{5}{c}{SFT Data} & \multirow{2}{*}{Acc.} \\ \cline{2-6}
\multicolumn{1}{l}{} & Data Source & Volume & Average Resolution & Domain & Tools & \\\midrule
\multicolumn{1}{l}{Qwen-2.5VL-7B} & SuperRSVQA~\citep{wang2025geollava8kscalingremotesensingmultimodal} & 1.3k & 8,376$\times$8,376 & Remote Sensing & None & 46.6 \\\midrule
\multirow{5}{*}{Qwen-2.5VL-7B} & SuperRSVQA~\citep{wang2025geollava8kscalingremotesensingmultimodal} & 1.3k & 8,376$\times$8,376 & Remote Sensing & None & \multirow{5}{*}{42.7\textcolor{red}{$\downarrow$}} \\
 & vqa\_cold\_start~\citep{wu2025reinforcingspatialreasoningvisionlanguage} & 10k & 478$\times$390 & General Domain & Object Mapper, Object Mapper & \\
 & maze\_cold\_start~\citep{wu2025reinforcingspatialreasoningvisionlanguage} & 7.3k & 588$\times$588 & General Domain & Object Mapper, Object Mapper & \\
 & GPT4Scene\_cold\_start~\citep{wu2025reinforcingspatialreasoningvisionlanguage} & 6.1k & 504$\times$364 & General Domain & Object Mapper, Object Mapper & \\
 & SR\_91k\_cold\_start~\citep{wu2025reinforcingspatialreasoningvisionlanguage} & 11.4k & 504$\times$364 & General Domain & Object Mapper, Object Mapper & \\ \bottomrule
\end{tabular}
}
\vspace{-2mm}
\end{table}

\begin{wrapfigure}[13]{r}{0.50\textwidth}
\vspace{-3mm}
\centering
\includegraphics[width=\linewidth]{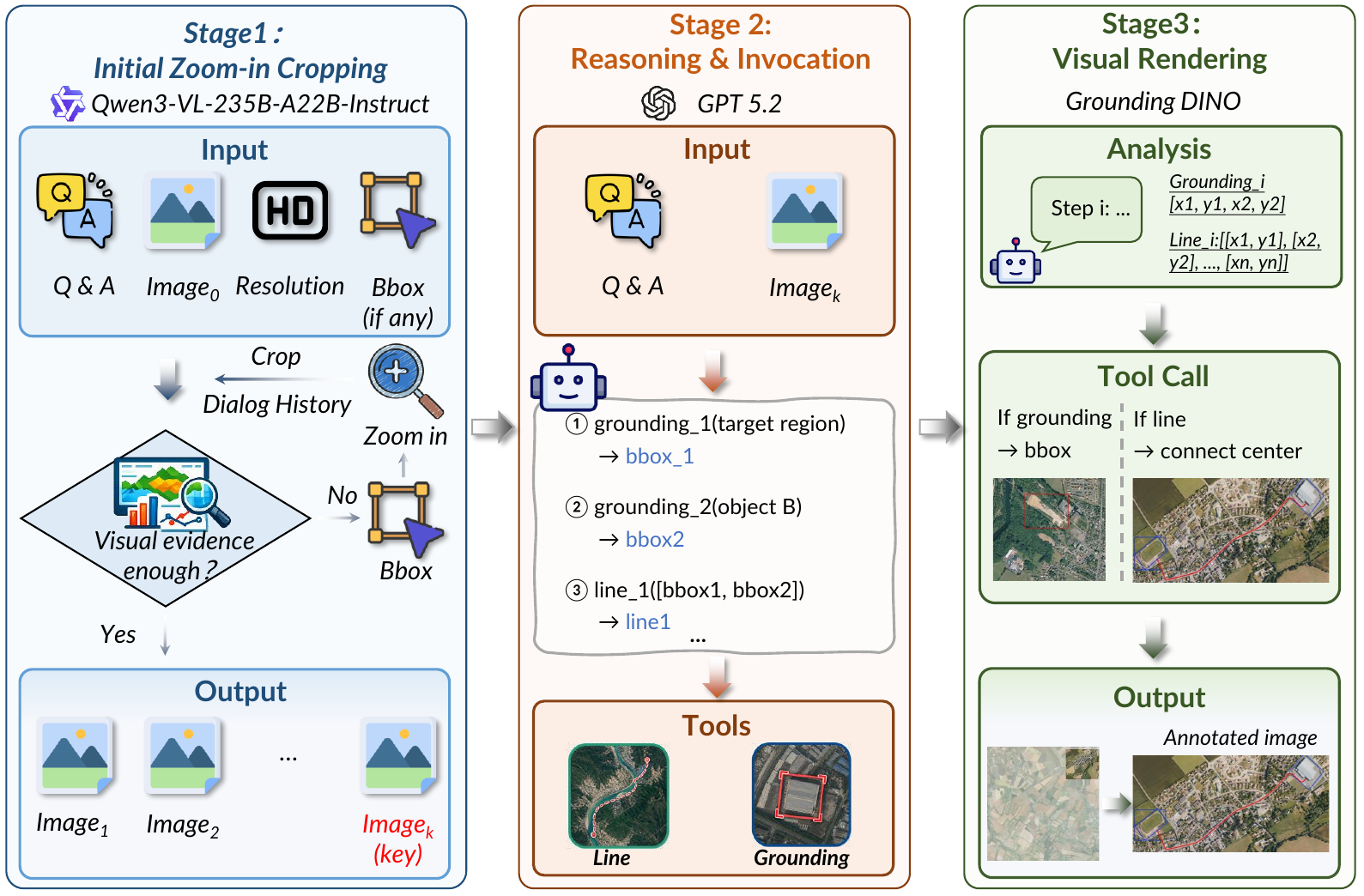}
\caption{\textbf{Data Construction Pipeline.}}
\label{fig:dataset-pipeline}
\vspace{-3mm}
\end{wrapfigure}

\noindent
\textbf{Pilot Experiments of Dataset.}
General-domain datasets have recently provided supervised traces for visual tool use, making it tempting to combine them with high-resolution RS VQA data for training tool-augmented RS MLLMs.
We test this strategy by only changing the mixture data in the SFT stage.
As shown in Table~1, Qwen2.5-VL-7B~\citep{qwen2.5} trained with SuperRSVQA~\citep{wang2025geollava8kscalingremotesensingmultimodal} alone achieves 46.6\% accuracy on XLRS-Bench~\citep{xlrs-bench}, whereas adding several general-domain tool-interaction datasets reduces the accuracy to 42.7\%.
This result suggests that simply increasing SFT data with visual tool-use trajectories does not improve UHR RS reasoning.

Most added datasets are built on low-resolution natural or synthetic images, whose visual scale, object distribution, and evidence layout differ substantially from UHR RS imagery.
Consequently, their tool-use traces may provide weak or even misleading supervision for RS scenarios, where task-relevant evidence is sparse, spatially dispersed, and often requires coordinated exploration across multiple regions.
These findings reveal a clear domain mismatch and motivate UHR RS-specific multi-tool interaction data.
\vspace{-6mm}
\begin{figure}[H]
\centering
\setlength{\tabcolsep}{0pt}

\newlength{\GeoLeftW}
\newlength{\GeoRightW}
\newlength{\GeoGapW}
\newlength{\GeoBodyH}
\newsavebox{\GeoTableBox}

\setlength{\GeoLeftW}{0.46\textwidth}
\setlength{\GeoGapW}{0.04\textwidth}
\setlength{\GeoRightW}{0.50\textwidth}

\sbox{\GeoTableBox}{%
\begin{minipage}{\GeoRightW}
\centering
\captionof{table}{\textbf{Main statistics of the GeoMTVR dataset.}}
\label{tab:GeoMTVR-stats}
\vspace{-1mm}
\small
\renewcommand{\arraystretch}{0.98}
\resizebox{\linewidth}{!}{
\begin{tabular}{lc}
\toprule
\textbf{Metric} & \textbf{Value} \\
\midrule
Total samples & 13,078 \\
Average image resolution & 6,033 $\times$ 6,017 \\
Average question length & 60.72 words \\
Average reasoning length & 126.92 words \\
Tool-used samples & 97.09\% \\
Average number of tool calls per sample & 1.53 \\
Average number of tool types per sample & 1.16 \\
\midrule
Samples containing zoomin & 78.36\% \\
Samples containing grounding & 39.64\% \\
Samples containing line & 5.31\% \\
\bottomrule
\end{tabular}
}
\end{minipage}
}

\setlength{\GeoBodyH}{\dimexpr\ht\GeoTableBox+\dp\GeoTableBox\relax}

\begin{tabular}{@{}p{\GeoLeftW}@{\hspace{\GeoGapW}}p{\GeoRightW}@{}}

\begin{minipage}[t][\GeoBodyH][t]{\linewidth}
\vspace{0pt}
\centering
\includegraphics[
  height=\GeoBodyH,
  keepaspectratio,
  trim={6mm 6mm 6mm 6mm},
  clip
]{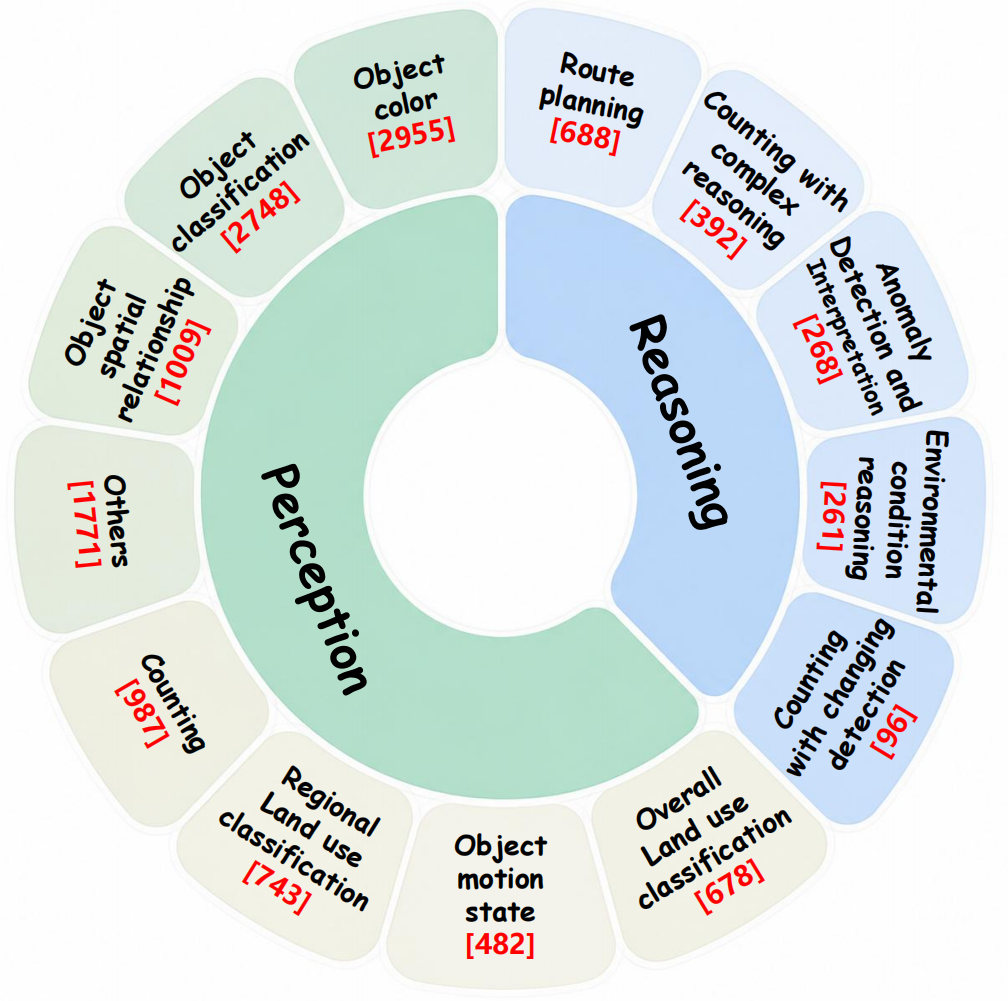}
\end{minipage}
&
\begin{minipage}[t][\GeoBodyH][t]{\linewidth}
\vspace{0pt}
\centering
\usebox{\GeoTableBox}
\end{minipage}
\\[-0.5mm]

\begin{minipage}[t]{\linewidth}
\centering
\captionsetup{type=figure,skip=3pt}
\captionof{figure}{\textbf{Our dataset covers perception and reasoning dimensions across 13 sub-tasks.}}
\label{fig:dataset-dimension}
\end{minipage}
&
\begin{minipage}[t]{\linewidth}
\centering
\end{minipage}

\end{tabular}

\vspace{-6mm}
\end{figure}

\textbf{Data source selection.} 
Our dataset is built on existing high resolution remote sensing VQA resources.
We integrate multi-source imagery and apply unified re-annotation to improve data diversity and training utility.
Specifically, we use high-resolution VQA sources including SuperRSVQA~\citep{wang2025geollava8kscalingremotesensingmultimodal} and LRS-GRO-train~\citep{liu2025zoomearth}. 
To ensure data quality and fair evaluation, we perform strict image-level deduplication and remove samples overlapping with existing benchmarks such as XLRS-Bench~\citep{xlrs-bench} and LRS-GRO-eval.



\textbf{Annotation.}
Directly using current MLLMs, such as GPT-5.4~\citep{openai_gpt5_4_2026} and Claude 4.6~\citep{anthropic_2026_claude_4_6}, to annotate tool-use behaviors in UHR satellite imagery often leads to unreliable results.
We observe pronounced hallucination and instability, especially in the initial question-region cropping stage, where localization shifts or target misidentification can cascade into incorrect bounding-box and auxiliary-line annotations.
To reduce such errors, we adopt the staged annotation framework shown in Figure~\ref{fig:dataset-pipeline}.
Given a question, image, and answer, we first use Qwen3-VL-235B-A22B~\citep{Qwen3-VL} to coarsely crop the question-relevant region.
We then incorporate VQA semantics and use GPT-5.2~\citep{openai_gpt5_2_2025} to generate tool-augmented reasoning traces, following the general idea of constructing detailed visual reasoning processes through chained visual manipulations~\citep{qi2024cogcom}.
Finally, GroundingDINO~\citep{liu2023grounding} is applied to precisely localize the objects involved in the question, enabling the progressive construction of structured visual reasoning annotations.

\textbf{Statistics.} 
Fig.~\ref{fig:dataset-dimension} visualizes the capability dimensional coverage of both datasets. It can be seen that our datasets span a wider range of task dimensions, showing strong alignment with real-world scenarios.
What's more, we summarize the key statistics of our datasets in Tab.~\ref{tab:GeoMTVR-stats}, including average question/answer lengths, the diversity and quantity of annotated objects, average resolution, and other relevant attributes.

\section{Method}

\subsection{
Pilot study: limitations of single zoom-in tool in MLLM.
}
Existing studies have incorporated tool invocation into high-resolution image VQA by interleaving textual reasoning with visual operations such as zoom-in~\citep{zheng2025deepeyes,hong2025deepeyesv2,shen2025zoomeye,su2025thinking,wang2025vicot}. Given an image and a question, the model decides whether to answer directly or acquire higher-resolution local evidence through visual actions.
Recent general-domain studies have questioned the broad effectiveness of zoom-in, arguing that its gains may mainly come from reinforcement learning on effective data rather than from the tool itself~\citep{hou2025codev,wei2026zooming}.
In contrast, RS UHR studies such as GeoEyes and ZoomEarth report clear benefits of zoom-in for ultra-high-resolution remote-sensing understanding.
\textbf{This discrepancy raises a key question: whether zoom-in is genuinely effective in RS UHR scenarios, where its limitations emerge, and which types of problems it can solve.}

To examine this question, we conduct a pilot study on XLRS-Bench using Qwen3-VL-8B-Instruct as the base model. The model is trained with GRPO and equipped with only a zoom-in tool~\citep{deepseekmath2024,deepseekr1_2025}. The training data include approximately 7K samples from the original MED training set~\citep{ma2026does} and 6K samples from SuperRSVQA, forming a mixed corpus for tool-use RL in UHR remote-sensing scenarios. We use a batch size of 256, set the rollout number to 8, and evaluate the model every 10 steps for 220 training steps. Following MED, we decompose tool-induced performance changes into \textbf{Call Gain, Schema Gain, Call Harm, and Schema Harm}, and further factorize Call Gain into \textbf{Mass, Policy, and Quality}. We focus on the Quality term, namely the probability that a tool call leads to a correct answer on failed samples.

\begin{wrapfigure}{r}{0.50\textwidth}
\vspace{-4mm}
\centering
\includegraphics[width=\linewidth]{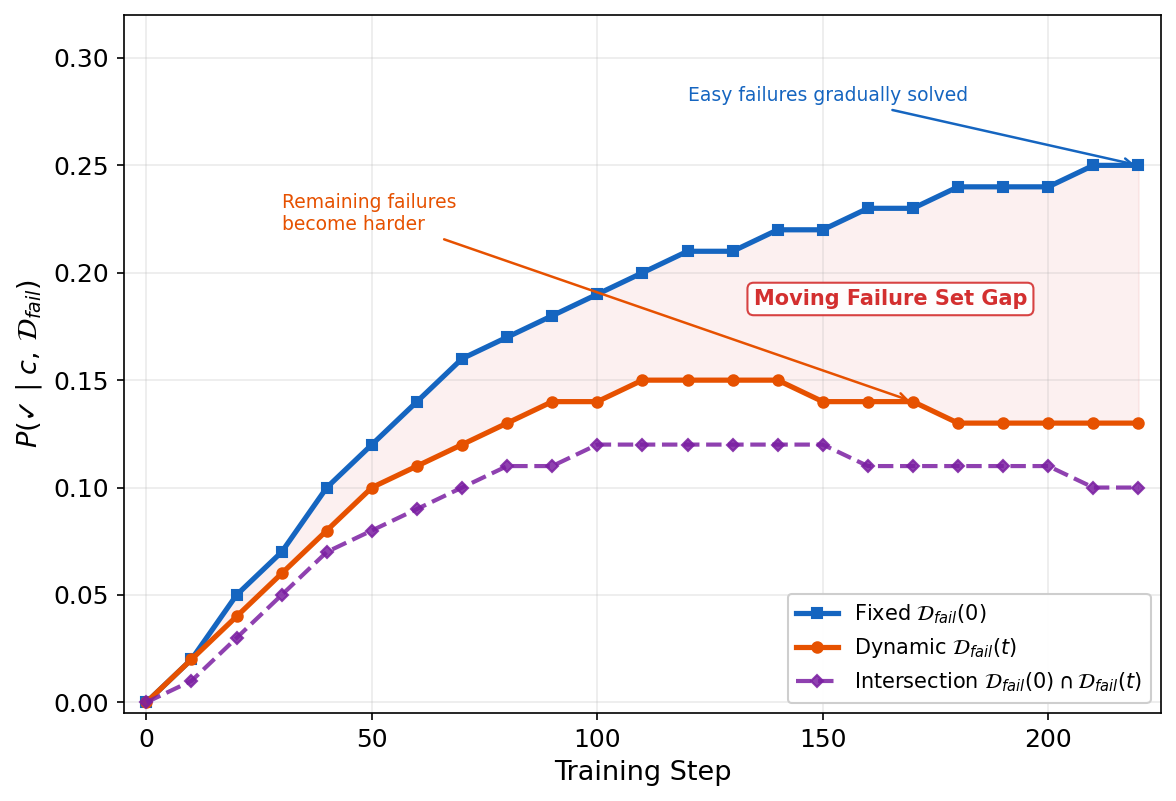}
\caption{\textbf{Quality factor under three failure-set definitions on XLRS-Bench.} The gap between fixed and dynamic failure sets reveals the moving failure set effect: as training progresses, easier failures are resolved, while the remaining dynamic failures require more complex evidence acquisition than a single zoom-in tool can provide.}
\vspace{-4mm}
\label{fig:moving_failure_set}
\end{wrapfigure}
We consider three failure sets: the dynamic failure set $\mathcal{D}_{\mathrm{fail}}(t)$, containing samples answered incorrectly without tools at checkpoint $t$; the fixed failure set $\mathcal{D}_{\mathrm{fail}}(0)$, defined by the base model before RL; and their intersection $\mathcal{D}_{\mathrm{fail}}(0)\cap\mathcal{D}_{\mathrm{fail}}(t)$, which captures persistently difficult samples. As shown in Fig.~\ref{fig:moving_failure_set}, the fixed Quality increases steadily from 0.00 to 0.25, indicating that RL learns useful zoom-in behaviors on the original failure set. However, the dynamic Quality first increases and then declines, peaking at around 0.15 and ending at 0.13. This divergence reveals a moving failure set effect: as easier samples are solved and leave $\mathcal{D}_{\mathrm{fail}}(t)$, the remaining failures become harder, and a single zoom-in operation can no longer provide sufficient evidence. By the final checkpoint, the fixed Quality is nearly twice the dynamic Quality, suggesting that evaluation on the initial failure set substantially overestimates tool effectiveness on current hard cases. Table~\ref{tab:zoomin_qualitative} further explains this effect.

\begin{AIbox}{Takeaway}
Zoom-in tool in MLLM mainly solves easy UHR RS tasks with localized evidence, such as object recognition, local counting, and attribute identification, but remains insufficient for hard tasks requiring global coverage, multi-region comparison, or long-range spatial reasoning.
\end{AIbox}
\vspace{-3mm}

Table~\ref{tab:zoomin_qualitative} further explains this effect. Solved samples usually depend on evidence from one localized region, such as recognizing a vehicle type, counting objects in a specified port area, identifying a roof color, or determining the relative position of nearby landmarks. In these cases, the main challenge is to locate the relevant region, and one zoom-in operation often provides enough detail. In contrast, persistently failed samples involve global counting, path planning, land-use enumeration, change counting across image pairs, or reasoning over spatially dispersed clues. These tasks require systematic spatial coverage, multi-region comparison, or heterogeneous perceptual capabilities beyond zoom-in. Although the framework allows up to five zoom-in calls and retains the global image in context, limited coverage and the absence of structured exploration still prevent the model from collecting sufficient evidence.

\begin{table*}[h]
\vspace{-3mm}
\centering
\small
\caption{\textbf{More details of Pilot Experiments.} Qualitative analysis of samples solved by single-tool zoom-in and those that remain persistently failed. Single zoom-in is effective for localized evidence acquisition, but struggles with global coverage, multi-region comparison, and structured spatial reasoning in UHR remote-sensing images.}
\label{tab:zoomin_qualitative}
\setlength{\tabcolsep}{4pt}
\renewcommand{\arraystretch}{1.12}
\begin{tabular}{p{0.16\linewidth} p{0.25\linewidth} p{0.50\linewidth}}
\toprule
\textbf{Group} & \textbf{Representative tasks} & \textbf{Key observation} \\
\midrule
\multicolumn{3}{l}{\textit{Samples solved during RL training}} \\
\midrule
Object classification 
& Vehicle-type recognition 
& A single zoom-in to the target region reveals discriminative visual details, such as the difference between trucks and buses. \\

Regional counting 
& Counting ships in a specified port 
& The relevant region is spatially localized, and objects become clearly separable after zoom-in. \\

Attribute recognition 
& Identifying roof color 
& Zoom-in mitigates the loss of fine-grained appearance caused by global-view downsampling. \\

Local spatial relation 
& Reasoning over nearby landmarks 
& Both targets can be covered by one local crop, making their relative layout directly observable. \\
\midrule
\multicolumn{3}{l}{\textit{Persistently failed samples}} \\
\midrule
Global counting 
& Counting vehicles in a city-scale image 
& Limited zoom-in calls cannot systematically cover the full UHR image, leaving many targets unobserved. \\

Path planning 
& Finding a route between distant points 
& The global view lacks road-level details, while local zoom-in crops are insufficient to form a coherent long-range path. \\

Land-use enumeration 
& Identifying all land-use types 
& The answer depends on multiple heterogeneous regions, but limited local inspections cover only a small subset of them. \\

Change counting 
& Comparing object counts across image pairs 
& Both images require multiple local inspections, making the fixed tool budget insufficient for reliable comparison. \\

Complex reasoning 
& Inferring conditions from dispersed clues 
& The required evidence is distributed across distant regions, and crop-and-zoom lacks a structured exploration mechanism. \\
\bottomrule
\end{tabular}
\vspace{-4mm}
\end{table*}

\subsection{GeoLens}
\label{subsec:geolens}
GeoLens is trained in two stages. We first perform supervised fine-tuning (SFT) on GeoMTVR to initialize multi-tool visual reasoning behaviors, enabling the model to imitate interleaved trajectories of reasoning, tool invocation, and visual observation. We then optimize the model with a tool-attention-focused reinforcement learning objective that strengthens attention routing around key tool-use decisions. This design separates behavior acquisition from policy refinement: SFT teaches the model how to use different tools, while reinforcement learning improves when and how they should be invoked.

\begin{figure*}[!t]
\centering
\includegraphics[width=0.98\textwidth]{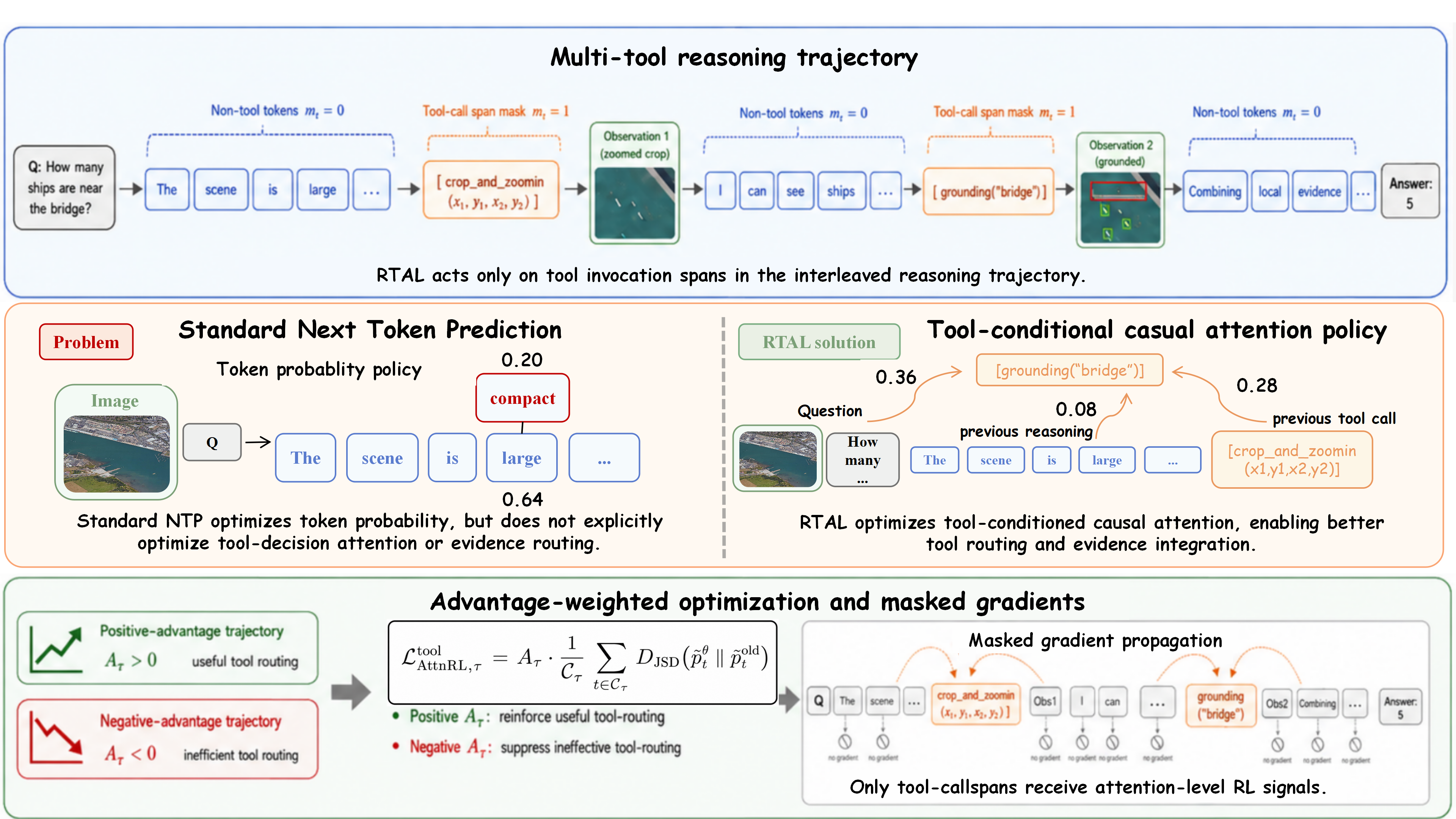}
\caption{\textbf{Overview of Reinforced Tool Attention Learning.} RTAL focuses reinforcement signals on tool-call spans, encouraging the model to route attention toward task-relevant context and tool observations during multi-tool visual reasoning.}
\label{fig:rtal-overview}
\vspace{-6mm}
\end{figure*}

\subsubsection{Reinforced Tool Attention Learning}
In tool-augmented UHR remote-sensing reasoning, the challenge is not only to produce the correct answer, but also to allocate computation to the right visual evidence acquisition process. A trajectory contains textual reasoning tokens, tool invocations, returned observations, and subsequent interpretation tokens. Standard policy-gradient methods optimize generated tokens under a scalar reward and thus treat most tokens with a similar objective. This is suboptimal for multi-tool reasoning, since decisive errors often occur around a small set of tool-related decisions: whether to call a tool, which tool to invoke, where to apply it, and how to use the returned observation. Uniform reinforcement over all tokens can dilute the learning signal for these decisions and lead to inefficient or homogeneous tool use.

Recent work such as RAL~\citep{li2026ral} shows that MLLM post-training can benefit from optimizing internal attention allocation rather than only next-token likelihood. However, existing attention-based RL mainly targets standard image or video QA, where the visual input remains fixed during generation. It does not explicitly handle agentic tool-use trajectories, where each tool call changes the available evidence and affects later reasoning and future tool choices. We therefore reformulate attention-based reinforcement learning for multi-tool visual reasoning. Instead of optimizing attention over all generated tokens, we focus on attention patterns associated with tool-call spans in the chain-of-thought process. By strengthening policy updates on these tool-related attention trajectories, our method encourages the model to condition later reasoning and tool invocations on previous tool outputs, yielding more adaptive and evidence-grounded multi-tool policies.

Formally, reinforced tool attention learning has three steps. First, we extract tool-call spans from the generated trajectory and define a tool-conditioned causal attention policy over their valid context. Second, we apply an advantage-weighted divergence objective to promote attention routing patterns from high-reward trajectories and suppress those from low-reward ones. Third, we use a time-step mask so that attention-level gradients are propagated only through tool-call subsequences. This preserves ordinary language generation while concentrating optimization on the causally sensitive regions of tool invocation.

Unless otherwise specified, $\theta$ denotes the current model parameters, $\tau$ a sampled reasoning trajectory, $D_{\text{JSD}}(\cdot\parallel\cdot)$ the Jensen--Shannon divergence, and $\mathbb{E}[\cdot]$ expectation over the indicated sampling process. We use $L_{\text{RL}}$ for the standard output-level RL objective and $L_{\text{AttnRL}}^{\text{tool}}$ for the proposed tool-focused attention objective.

\textbf{Tool-Conditioned Causal Attention Policy.}
In our implementation, reinforcement attention learning is applied only to generation subsequences related to tool calls, compressing the original full-sequence attention policy into a tool-triggered conditional subspace. Let the full sequence be $S=(x_1,\ldots,x_T)$, where $x_{1:P}$ is the prompt and $x_{P+1:T}$ is the generated tokens. Define
\begin{equation}
\label{eq:tool-token-set}
\mathcal{C}=\{t \mid x_t \in \texttt{ToolsCall}\}
\end{equation}
as the set of time steps inside tool-call spans (including function names, argument key-value pairs, and structural markers). For any $t\in\mathcal{C}$, let $\alpha_{t,i}$ ($i<t$) denote attention weights from the final layer. To characterize information routing driven only by tool calls, we normalize attention over valid causal context and obtain the conditional attention policy:
\begin{equation}
\label{eq:tool-conditioned-attention-policy}
\tilde{p}_t^\theta(i)=\frac{\alpha_{t,i}}{\sum_{j=1}^{t-1}\alpha_{t,j}},\quad
i\in\{1,\ldots,t-1\},\ t\in\mathcal{C},
\end{equation}
where $T$ is the sequence length, $P$ is the prompt length, $x_t$ is the token at time step $t$, $\mathcal{C}$ is the set of tool-call time steps, $i$ indexes a previous context token, $\alpha_{t,i}$ is the averaged final-layer attention weight from token $t$ to token $i$, $\tilde{p}_t^\theta(i)$ is the normalized tool-conditioned attention probability under current parameters $\theta$. This form is consistent with standard definitions, but optimization is performed only on $\mathcal{C}$, explicitly constraining policy semantics to how the model integrates prompt and generated reasoning to make executable structured decisions during tool calls, without imposing unnecessary constraints on non-tool text spans.

\textbf{Advantage-Weighted Divergence over Tool Tokens.}
To realize reward-based policy improvement in the tool-call subspace, we rewrite the advantage-weighted divergence objective from a full-sequence expectation to a conditional expectation over $\mathcal{C}$, applying optimization signals only at tool-related steps. Let $\tilde{p}_t^{\text{old}}$ be the old policy, and let $A_\tau$ be sequence-level or segment-level advantage (shared over $\mathcal{C}$ in practice). Using symmetric and bounded Jensen--Shannon divergence for numerical stability, the attention reinforcement objective is
\begin{equation}
\label{eq:tool-attention-rl-loss}
L_{\text{AttnRL}, \tau}^{\text{tool}}
=
A_\tau\cdot
\frac{1}{\mathcal{C}_\tau}\sum_{t\in\mathcal{C}_{\tau}}
D_{\text{JSD}}\!\left(
\tilde{p}_t^\theta \parallel \tilde{p}_t^{\text{old}}
\right)
.
\end{equation}
When minimizing this loss, a positive advantage $A_\tau>0$ pulls the current tool attention toward successful historical routing, while a negative advantage $A_\tau<0$ pushes attention away from inefficient routing patterns. Because the expectation domain is restricted to $\mathcal{C}$, gradients are not diluted across long non-critical spans, strengthening supervision for structured argument selection, field alignment, and cross-step dependency modeling. The final objective is linearly coupled with standard policy gradient:
\begin{equation}
\label{eq:total-rtal-loss}
L_{\text{total}}
=
L_{\text{RL}}
+\lambda_{\text{attn}}\,L_{\text{AttnRL}}^{\text{tool}},
\end{equation}
where $\tilde{p}_t^{\text{old}}$ is the frozen attention policy before the current update, $A_\tau$ is the advantage of trajectory $\tau$, $L_{\text{total}}$ is the final training loss, and $\lambda_{\text{attn}}$ balances output-level optimization and internal routing exploration, with effective scope limited to tool-call spans.

\textbf{Masked Gradient Propagation for Tool-Specific Attention.}
At the gradient level, we apply an explicit time-step mask to backpropagate only within tool-call spans, precisely projecting distribution-level gradients onto corresponding attention logits. Let
\begin{equation}
\label{eq:jsd-local-objective}
J_t=D_{\text{JSD}}\!\left(\tilde{p}_t^\theta\parallel\tilde{p}_t^{\text{old}}\right).
\end{equation}
where $J_t$ is the local divergence objective at tool-call step $t$. Its distribution gradient $\nabla_{\tilde{p}_t^\theta}J_t$ follows the closed form of JSD. Let $e_{t,i}$ denote the attention logit before softmax normalization and use $\tilde{p}_i$ as shorthand for $\tilde{p}_t^\theta(i)$ at a fixed $t$. Using the softmax Jacobian
\begin{equation}
\label{eq:softmax-jacobian}
\frac{\partial \tilde{p}_i}{\partial e_j}
=
\tilde{p}_i(\delta_{ij}-\tilde{p}_j),
\end{equation}
the chain rule gives the logit gradient:
\begin{equation}
\label{eq:attention-logit-gradient}
\nabla_{e_{t,i}}J_t
=
\tilde{p}_t^\theta(i)
\left(
\nabla_{\tilde{p}_t^\theta(i)}J_t
-\sum_j \tilde{p}_t^\theta(j)\nabla_{\tilde{p}_t^\theta(j)}J_t
\right).
\end{equation}
The parameter gradient becomes
\begin{equation}
\label{eq:masked-rtal-gradient}
\nabla_\theta L_{\text{AttnRL}}^{\text{tool}}
=
\mathbb{E}_{\tau}
\left[
A_\tau
\sum_{t=1}^{T}
\sum_{i=1}^{t-1}
\nabla_{e_{t,i}}J_t\cdot
\frac{\partial e_{t,i}}{\partial\theta}
\right].
\end{equation}
Here, $\delta_{ij}$ is the Kronecker delta, $j$ indexes context positions in the softmax Jacobian and $\partial e_{t,i}/\partial\theta$ maps the logit-level gradient back to model parameters. In practice, $L_{\text{AttnRL}}^{\text{tool}}$ is averaged over all attention heads in the final layer. In equation (~\ref{eq:masked-rtal-gradient}), only attention paths inside tool-call subsequences receive reinforcement signals. This masked backpropagation locally reshapes internal information routing without altering the forward graph, allowing the model to preserve linguistic generation flexibility while learning reward-aligned attention allocation and cross-step dependency structures in the causally sensitive region of tool invocation. 

\vspace{-2mm}
\section{Experiments}
\label{section5}

\textbf{Experimental Setup}
We first conduct supervised training with Qwen2.5-VL-7B on our newly annotated GeoMTVR dataset, using interleaved image-text supervision and explicit tool-call annotations. Specifically, we model three explicit tool types: \texttt{crop\_and\_zoomin}, \texttt{grounding}, and \texttt{line}. In the reinforcement stage, based on the our RTAL method, we perform reinforcement learning on cleaned SuperRS-VQA and LRS-GRO-train data. For evaluation, we test on XLRS-Bench and LRS-GRO-eval. \textbf{Additional details, extended ablations are provided in the Appendix.}

\begin{table}[htbp]
\footnotesize
\vspace{-0.4cm}
\caption{
\textbf{Experimental results on the perception and reasoning dimensions on XLRS-Bench, with models ranked by average performance.}  We mark the highest score in \colorbox{backred!60}{red}.
}
\label{tab:vqa}
\centering
\resizebox{\textwidth}{!}{%
\begin{tabular}{l|cccccccc|ccccc|c}
\toprule \Gray
\multicolumn{1}{c}{\textbf{Method}} & \multicolumn{8}{c}{\textbf{Perception}}  & \multicolumn{5}{c}{\textbf{Reasoning}} &  \\  \midrule
\Gray\textbf{Sub-tasks (L-3 Capability)}  &\textbf{OC} & \textbf{RC} & \textbf{OLUC} & \textbf{RLUC}  & \textbf{OCC} & \textbf{OCL} & \textbf{OMS} & \textbf{OSR} & \textbf{AD}  & \textbf{ECR} & \textbf{RP} & \textbf{RCCD} & \textbf{CCR} & \textbf{Avg.} \\ \midrule
\multicolumn{15}{l}{\textit{\textcolor{gray}{Remote Sensing MLLMs}}} \\
GeoChat~\cite{geochat} & 16.7 & 29.0 & 2.0 & 23.0 & 21.1 & 16.8 & 35.0 & 24.2 & 33.0 & 43.0 & 10.0 & - & 21.0 & 22.9\\
ZoomEarth~\cite{liu2025zoomearth} & 30.0 & 38.0 & 8.0 & 63.0 & 35.1 & 33.9 & 65.0 & 31.4 & 66.0 & 73.0 & 22.0 & 28.3 & 40.0 & 41.1\\
GeoLLaVA-8K~\cite{wang2025geollava8kscalingremotesensingmultimodal} & 26.7 & 38.0 & \colorbox{backred!60}{49.0} & 69.0 & 41.6 & 31.6 & 65.0 & 35.0 & 67.0 & 78.0 & \colorbox{backred!60}{66.0} & 50.0 & 52.0 & 51.5\\
\midrule
\multicolumn{15}{l}{\textit{\textcolor{gray}{Closed-source MLLMs}}} \\ 
\Lgray Claude 3.7 Sonnet~\cite{anthropic_claude37_2025} & 27.6 & 22.7 & 17.4 & 68.4 & 30.5 & 29.9 & 63.6 & 27.6 & 64.8 & 78.4 & 34.5 & 27.8 & 32.6 & 40.5\\
\Lgray Gemini 2.5 Pro~\cite{gemini25pro_2025} & -- & -- & -- & -- & -- & -- & -- & -- & -- & -- & -- & -- & -- & 45.2\\
\Lgray GPT-5.2~\cite{openai_gpt5_2_2025} & 30.0 & 37.0 & 17.0 & 70.5 & 43.0 & 41.4 & \colorbox{backred!60}{68.3} & 34.0 & 74.0 & 76.0 & 52.0 & 36.7 & 38.0 & 47.5\\
\midrule
\multicolumn{15}{l}{\textit{\textcolor{gray}{Open-source MLLMs}}} \\
ViLaSR~\cite{wu2025reinforcingspatialreasoningvisionlanguage} & 28.3 & 38.0 & 17.0 & 72.0 & 31.1 & 33.1 & 65.0 & 39.4 & 70.0 & 76.0 & 27.0 & 41.7 & 47.0 & 45.1\\
InternVL3-8B~\cite{internvl3} & 40.0 & 39.0 & 10.0 & 71.5 & 44.5 & 30.8 & 65.0 & 25.2 & \colorbox{backred!60}{77.0} & 82.0 & 36.0 & 21.7 & 50.0 & 45.6\\
Qwen2-VL-7B~\cite{qwen2} & 26.7 & 40.0 & 11.0 & 73.0 & 35.9 & 34.6 & 61.7 & 31.8 & 70.0 & 81.0 & 35.0 & 46.7 & 48.0 & 45.8\\
InternVL2.5-8B~\cite{internvl25_2024} & 38.3 & 37.0 & 10.0 & 77.0 & 33.4 & 35.5 & 65.0 & 21.6 & 73.0 & \colorbox{backred!60}{83.0} & 34.0 & 50.0 & 43.0 & 46.2\\
Qwen2.5-VL-7B~\cite{qwen2.5} & 33.3 & 40.0 & 31.0 & 77.0 & 40.6 & 40.5 & 66.7 & 36.2 & 68.0 & 72.0 & 27.0 & 38.3 & 45.0 & 47.4\\
InternVL3-78B~\cite{internvl3} & 23.3 & 49.0 & 33.0 & 74.0 & 42.5 & 37.4 & 66.7 & 30.0 & 76.0 & 81.0 & 40.0 & 45.0 & 42.0 & 49.2\\
Qwen3-VL-8B~\cite{Qwen3-VL} & 21.7 & 50.0 & 26.0 & \colorbox{backred!60}{81.5} & 46.6 & \colorbox{backred!60}{43.1} & 66.7 & 30.4 & 74.0 & 79.0 & 37.0 & 43.3 & 51.0 & 50.0\\
Qwen2.5-VL-72B~\cite{qwen2.5} & 33.3 & 47.0 & 39.0 & 80.0 & 45.3 & 42.1 & 65.0 & 34.0 & 71.0 & 74.0 & 37.0 & 43.3 & 42.0 & 50.2\\
Qwen3-VL-235B-A22B~\cite{Qwen3-VL} & \colorbox{backred!60}{61.7} & \colorbox{backred!60}{82.5} & 38.6 & 36.7 & 44.0 & 39.0 & 38.9 & \colorbox{backred!60}{49.0} & 73.0 & 82.0 & 37.8 & 33.3 & 48.0 & 51.1\\
Intern-S1-mini~\cite{interns1mini_2024} & -- & -- & -- & -- & -- & -- & -- & -- & -- & -- & -- & -- & -- & 51.6\\
\midrule
\textbf{GeoLens (OURS)} & 35.0 & 40.0 & 37.0 & 76.5 & 46.5 & \colorbox{backred!60}{45.6} & 65.0 & 36.8 & 72.0 & 82.0 & 62.0 & \colorbox{backred!60}{51.7} & \colorbox{backred!60}{55.0} & \colorbox{backred!60}{54.2}\\

 \bottomrule
\end{tabular}%
}
\vspace{-4mm}
\end{table}

\begin{table}[htbp]
\footnotesize
\vspace{-2mm}
\caption{
\textbf{Detailed results on RSHR-Bench, with models ranked by average performance within each category.} We mark the highest score in \textcolor{red}{red}.
}
\label{tab:xlrs-detailed}
\centering
\resizebox{\textwidth}{!}{%
\begin{tabular}{l|ccccccccc|c|ccccc|c|ccc|c}
\toprule \Gray
\multicolumn{1}{c}{\textbf{Model}} & \multicolumn{10}{c}{\textbf{Perception}} & \multicolumn{6}{c}{\textbf{Reasoning}} & \multicolumn{3}{c}{\textbf{Advanced}} & \textbf{Overall} \\
\Gray
\textbf{Metric} & \textbf{COL} & \textbf{SHP} & \textbf{DET} & \textbf{OC} & \textbf{REL} & \textbf{OGD} & \textbf{RG} & \textbf{OCN} & \textbf{RCN} & \textbf{Avg.} & \textbf{AD} & \textbf{FP} & \textbf{MRJC} & \textbf{MRJCS} & \textbf{OSJ} & \textbf{Avg.} & \textbf{MAD} & \textbf{MTFP} & \textbf{MOSJ} & \textbf{Avg.} \\
\midrule
\multicolumn{21}{l}{\textit{\textcolor{gray}{Remote Sensing VLMs}}} \\
\rowcolor{gray!15} EarthDial~\cite{soni2024earthdial} & 41.0 & 22.0 & 21.0 & 30.0 & 32.5 & 30.5 & 27.1 & 18.0 & 31.0 & 28.1 & 42.0 & 30.0 & 29.5 & 32.0 & 52.0 & 37.1 & 56.7 & 60.0 & 73.5 & 37.0 \\
GeoChat~\cite{geochat} & 32.5 & 22.0 & 24.0 & 29.5 & 40.0 & 25.0 & 22.9 & 22.5 & 29.0 & 25.9 & 30.0 & 24.0 & 25.5 & 30.0 & 32.0 & 28.3 & 48.3 & 46.0 & 62.9 & 32.1 \\
VHM~\cite{pang2025vhm} & 25.5 & 25.0 & 26.0 & 26.5 & \textcolor{red}{55.0} & 25.0 & 22.9 & 25.0 & 25.0 & 25.7 & 26.0 & 24.0 & 26.5 & 34.0 & 28.0 & 27.7 & 45.0 & 53.3 & 46.2 & 31.7 \\
\rowcolor{gray!15} GeoLLaVA-8K~\cite{wang2025geollava8kscalingremotesensingmultimodal} & 25.0 & 24.0 & 25.0 & 25.0 & 25.0 & 25.0 & 21.4 & 25.0 & 25.0 & 24.5 & 24.0 & 0.0 & 0.0 & 34.0 & 22.0 & 16.0 & 25.0 & 24.7 & 47.7 & 23.4 \\
\midrule
\multicolumn{21}{l}{\textit{\textcolor{gray}{Open-source VLMs}}} \\
\rowcolor{gray!15} VILA-HD~\cite{shi2025scaling} & 40.0 & 22.0 & 22.0 & 37.0 & 35.5 & 26.0 & 21.4 & 24.5 & 24.0 & 28.0 & 58.0 & 30.0 & \textcolor{red}{55.0} & 32.0 & 58.0 & 46.6 & 65.0 & 57.3 & 57.6 & 39.1 \\
MiniCPM2 6~\cite{yao2024minicpm} & 21.5 & \textcolor{red}{28.0} & 30.0 & 24.0 & 19.5 & 29.5 & 34.3 & 22.0 & 29.0 & 27.4 & 26.0 & 30.0 & 35.0 & 32.0 & 30.0 & 30.6 & 26.7 & 23.3 & 31.1 & 27.8 \\
InternVL 3.5 8B~\cite{wang2025internvl3_5} & 21.5 & \textcolor{red}{28.0} & 18.0 & 21.5 & 29.0 & 28.5 & 30.0 & \textcolor{red}{26.5} & 25.0 & 25.3 & 20.0 & 16.0 & 29.0 & 34.0 & 26.0 & 25.0 & 30.0 & 22.7 & 40.2 & 26.2 \\
\rowcolor{gray!15} Phi-3.5-Vision~\cite{microsoft2024phi3} & 25.0 & 24.0 & 25.0 & 25.0 & 23.5 & 25.0 & 22.9 & 25.0 & 25.0 & 24.5 & 24.0 & 22.0 & 23.5 & 30.0 & 22.0 & 24.3 & 28.3 & 24.7 & 47.0 & 26.0 \\
Deepseek-VL~\cite{deepseek-vl} & 22.5 & 22.0 & 21.0 & 25.0 & 20.5 & 26.0 & 28.6 & 20.5 & 22.0 & 23.1 & 22.0 & 28.0 & 50.0 & 32.0 & 20.0 & 30.4 & 20.0 & 23.3 & 33.3 & 25.7 \\
InternVL2.5-8B~\cite{internvl25_2024} & 25.5 & 22.0 & 26.0 & 26.0 & 22.5 & 24.5 & 30.0 & 22.5 & 20.0 & 24.3 & 26.0 & 20.0 & 22.5 & 34.0 & 20.0 & 24.5 & 25.0 & 28.7 & 35.6 & 25.3 \\
Qwen2.5-VL-7B~\cite{qwen2.5} & 29.5 & 25.0 & 22.0 & 28.0 & 25.0 & 24.5 & 24.3 & \textcolor{red}{26.5} & 22.0 & 25.2 & 26.0 & 28.0 & 25.0 & 10.0 & 20.0 & 21.8 & 21.7 & 24.0 & 10.6 & 23.1 \\
\midrule
\multicolumn{21}{l}{\textit{\textcolor{gray}{Closed-source VLMs}}} \\
\rowcolor{gray!15} GPT-4o~\cite{gpt4o} & 49.5 & 23.0 & 15.0 & 35.5 & 30.5 & 28.0 & 27.1 & 22.5 & \textcolor{red}{41.0} & 30.2 & 68.0 & 56.0 & 30.5 & 32.0 & 64.0 & 50.1 & 70.0 & 72.0 & 84.1 & 44.0 \\
GPT5~\cite{openai_gpt5_4_2026} & 29.0 & 10.0 & 23.0 & 23.0 & 37.0 & 24.5 & 31.4 & 20.0 & 23.0 & 24.5 & \textcolor{red}{74.0} & \textcolor{red}{58.0} & 35.0 & 34.0 & \textcolor{red}{66.0} & \textcolor{red}{53.4} & \textcolor{red}{78.3} & \textcolor{red}{73.3} & \textcolor{red}{86.4} & 42.7 \\
GPT-4o-mini~\cite{gpt4omini} & 41.5 & 16.0 & 29.0 & 31.5 & 31.5 & 32.0 & 28.6 & 19.5 & 32.0 & 29.1 & 54.0 & 54.0 & 31.5 & \textcolor{red}{48.0} & 54.0 & 48.3 & \textcolor{red}{78.3} & 68.0 & 75.0 & 42.6 \\
\rowcolor{gray!15} Gemini-2.5-pro~\cite{gemini25pro_2025} & 55.0 & 18.0 & \textcolor{red}{31.0} & \textcolor{red}{40.0} & 41.5 & 32.5 & \textcolor{red}{45.7} & 25.0 & 25.0 & \textcolor{red}{34.9} & 66.0 & 32.0 & 41.5 & 38.0 & 50.0 & 45.5 & 56.7 & 60.0 & 57.6 & 42.1 \\
\midrule
\multicolumn{21}{l}{\textit{\textcolor{gray}{OURS}}} \\
\rowcolor{gray!15} \textbf{GeoLens (OURS)} & \textcolor{red}{58.5} & 22.7 & 27.0 & 35.0 & \textcolor{red}{57.5} & \textcolor{red}{33.0} & \textcolor{red}{38.6} & \textcolor{red}{34.5} & 39.0 & \textcolor{red}{38.4} & 68.0 & 46.0 & 50.0 & 42.0 & 62.0 & \textcolor{red}{53.6} & 70.0 & 61.1 & 85.4 & \textcolor{red}{48.8} \\
\bottomrule
\end{tabular}%
}
\vspace{-4mm}
\end{table}

\begin{table}[htbp]
\footnotesize
\caption{\textbf{Results on the LRS-GRO-eval benchmark.} }
\label{tab:leaderboard}
\centering
\resizebox{0.82\textwidth}{!}{%
\begin{tabular}{l|l|c|c|c|c|c}
\toprule \Gray
\textbf{Leaderboard} & \textbf{LLM} & \textbf{MaxSize} & \textbf{Global} & \textbf{Region} & \textbf{Object} & \textbf{Avg} \\
\midrule
\rowcolor{gray!15} LLaVA-OV-1.5~\cite{llava_ov15_2025} & Qwen3-7B~\cite{qwen3} & 2304$\times$2304 & 62.43 & 36.18 & 39.92 & 45.33 \\
IXC-2.5~\cite{ixc2.5} & InternLM2-7B~\cite{internlm2_2023} & 4096$\times$4096 & 62.43 & 40.00 & 47.32 & 50.00 \\
InternVL3~\cite{internvl3} & InternLM3-8B~\cite{internlm3_2024} & 3200$\times$3200 & 71.60 & 44.58 & 47.80 & 53.67 \\
\rowcolor{gray!15} GeoChat~\cite{geochat} & Vicuna-1.5-7B~\cite{vicuna15_2023} & 504$\times$504 & 62.43 & 36.79 & 40.72 & 45.88 \\
Qwen2.5-VL (7B)~\cite{qwen2.5} & Qwen2.5-7B~\cite{qwen2.5} & 1024$\times$1024 & 69.39 & 38.47 & 43.58 & 49.62 \\
Qwen2.5-VL (3B)~\cite{qwen2.5} & Qwen2.5-3B~\cite{qwen2.5} & 1024$\times$1024 & 59.01 & 31.91 & 37.46 & 42.25 \\
\rowcolor{gray!15} VLM-R$^3$ (w/ tools)~\cite{jiang2025vlmr3} & Qwen3-7B~\cite{qwen3} & 512$\times$512 & 69.72 & 44.83 & 37.40 & 50.17 \\
GeoChat*~\cite{geochat} & Vicuna-1.5-7B~\cite{vicuna15_2023} & 504$\times$504 & 58.78 & 37.25 & 42.83 & 46.09 \\
ZoomEarth~\cite{liu2025zoomearth} & Qwen2.5-3B~\cite{qwen2.5} & 512$\times$512 & 63.09 & 46.11 & 51.80 & 53.76 \\
\midrule
\rowcolor{gray!15} \textbf{GeoLens (OURS)} & Qwen2.5-7B~\cite{qwen2.5} & 1024$\times$1024 & \colorbox{backred!60}{79.0} & \colorbox{backred!60}{48.9} & \colorbox{backred!60}{57.6} & \colorbox{backred!60}{60.7} \\

\bottomrule
\end{tabular}%
}
\vspace{-3mm}
\end{table}

\textbf{Main Results.}
As shown in Table~\ref{tab:vqa}, Table~\ref{tab:xlrs-detailed}, and Table~\ref{tab:leaderboard}, we conduct systematic evaluation on XLRS-Bench, RSHR-Bench, and LRS-GRO-eval. Our method achieves state-of-the-art performance with an average accuracy of 54.2\% on XLRS-Bench, and also reaches the best performance of 60.7\% on LRS-GRO-eval, consistently outperforming existing baselines. On the newly introduced fine-grained RSHR-Bench evaluation, GeoLens obtains the best overall average score of 48.8, ranking first among all compared models and showing clear advantages on challenging perception and reasoning dimensions such as COL and OGD.
Notably, with a 7B backbone, our model surpasses much larger models such as Qwen3-VL-235B (51.1\%) and Qwen2.5-VL-72B (50.2\%) on XLRS-Bench. It also achieves SOTA on LRS-GRO-eval and ranks first across all reported metrics. These results indicate that compared with paradigms without tool calls or with only a single tool, multi-tool calling significantly improves understanding and reasoning in ultra-high-resolution remote-sensing scenarios, highlighting its importance and future potential for complex vision-language analysis.

\section{Conclusion}
\vspace{-2mm}

In this paper, we addressed the challenge of visual evidence acquisition in ultra-high-resolution remote-sensing reasoning. Through a pilot study on XLRS-Bench, we showed that single-tool zoom-in is useful for locally recoverable evidence but remains insufficient for harder cases requiring global search, multi-region comparison, and structured spatial reasoning. To tackle this issue, we introduced GeoMTVR, a UHR RS multi-tool visual reasoning dataset with interleaved reasoning trajectories, visual tool calls, and returned visual observations, covering crop-and-zoom, grounding, and auxiliary-line operations. We further proposed Reinforced Tool Attention Learning (RTAL), which focuses reinforcement optimization on critical tool-use decisions rather than uniformly optimizing all generated tokens. Building on GeoMTVR and RTAL, we developed GeoLens, a multi-tool visual reasoning MLLM for UHR remote sensing. GeoLens achieves state-of-the-art performance on both XLRS-Bench and LRS-GRO-eval, outperforming direct reasoning and single-tool baselines while producing more effective tool-use trajectories. These results suggest that UHR RS reasoning benefits not only from larger models, but also from domain-specific interaction data and targeted optimization of visual tool use. 
\textbf{Limitation.}  We have primarily focused on optical satellite imagery; evaluating GeoLens on other sensor modalities (e.g., synthetic aperture radar or multispectral images) will be important to ensure broader applicability. 


\bibliographystyle{unsrt}
\bibliography{references.bib}


\newpage
\appendix
\definecolor{indianred}{rgb}{0.8, 0.36, 0.36}
\definecolor{bleudefrance}{rgb}{0.19, 0.55, 0.91}
\definecolor{forestgreen}{rgb}{0.0, 0.5, 0.0}
\definecolor{ashgrey}{rgb}{0.7, 0.75, 0.71}
\definecolor{darkorange}{rgb}{1.0, 0.55, 0.0}
\definecolor{darkorchid}{rgb}{0.6, 0.2, 0.8}
\definecolor{BurntOrange}{rgb}{0.8, 0.33, 0.0}
\definecolor{mycolor_green}{HTML}{E6F8E0}
\definecolor{backred}{RGB}{255, 190, 190}
\definecolor{red}{RGB}{139, 0, 0}
\definecolor{purple}{HTML}{E6F8E0}
\definecolor{indianred}{rgb}{0.8, 0.36, 0.36}
\definecolor{bleudefrance}{rgb}{0.19, 0.55, 0.91}
\definecolor{forestgreen}{rgb}{0.0, 0.5, 0.0}
\definecolor{ashgrey}{rgb}{0.7, 0.75, 0.71}
\definecolor{darkorange}{rgb}{1.0, 0.55, 0.0}
\definecolor{darkorchid}{rgb}{0.6, 0.2, 0.8}
\definecolor{BurntOrange}{rgb}{0.8, 0.33, 0.0}
\definecolor{mycolor_green}{HTML}{E6F8E0}
\definecolor{backred}{RGB}{255, 190, 190}
\definecolor{red}{RGB}{139, 0, 0}
\definecolor{purple}{HTML}{E6F8E0}
\definecolor{verylightgray}{HTML}{E6F8E0}
\definecolor{lightgray}{gray}{0.95}
\definecolor{mycolor}{RGB}{128, 0, 255}
\definecolor{wfx}{RGB}{128, 0, 255}
\definecolor{wdcolor}{RGB}{128, 0, 255}
\definecolor{wdqcolor}{RGB}{255, 0, 0}
\definecolor{verylightgray}{HTML}{E6F8E0} 
\definecolor{lightgray}{gray}{0.95} 

\definecolor{mycolor}{RGB}{128, 0, 255}
\definecolor{wfx}{RGB}{128, 0, 255}

\newcommand{\yf}[1]{{\color{blue}yf: #1}}

\newcommand{\roundedboxpink}[1]{
  \tikz[baseline=(char.base)]{
    \node[anchor=south west, rounded corners, text height=1.5ex, text depth=.25ex, fill=pink, draw=none, text=black, font=\bfseries] (char) {#1};
  }
}
\newcommand{\roundedboxgreen}[1]{
  \tikz[baseline=(char.base)]{
    \node[anchor=south west, rounded corners, text height=1.5ex, text depth=.25ex, fill=green!30, draw=none, text=black, font=\bfseries] (char) {#1};
  }
}
\newcommand{\roundedboxblue}[1]{
  \tikz[baseline=(char.base)]{
    \node[anchor=south west, rounded corners, text height=1.5ex, text depth=.25ex, fill=blue!30, draw=none, text=black, font=\bfseries] (char) {#1};
  }
}
\newcommand{\roundedboxyellow}[1]{
  \tikz[baseline=(char.base)]{
    \node[anchor=south west, rounded corners, text height=1.5ex, text depth=.25ex, fill=yellow!50, draw=none, text=black, font=\bfseries] (char) {#1};
  }
}
\newcommand{\roundedboxred}[1]{
  \tikz[baseline=(char.base)]{
    \node[anchor=south west, rounded corners, text height=1.5ex, text depth=.25ex, fill=red!30, draw=none, text=black, font=\bfseries] (char) {#1};
  }
}
\newcommand{\roundedboxpurple}[1]{
  \tikz[baseline=(char.base)]{
    \node[anchor=south west, rounded corners, text height=1.5ex, text depth=.25ex, fill=purple!50, draw=none, text=black, font=\bfseries] (char) {#1};
  }
}
\newcommand{\roundedboxbrown}[1]{
  \tikz[baseline=(char.base)]{
    \node[anchor=south west, rounded corners, text height=1.5ex, text depth=.25ex, fill=brown!30, draw=none, text=black, font=\bfseries] (char) {#1};
  }
}
\newcommand{\roundedboxorange}[1]{
  \tikz[baseline=(char.base)]{
    \node[anchor=south west, rounded corners, text height=1.5ex, text depth=.25ex, fill=orange!30, draw=none, text=black, font=\bfseries] (char) {#1};
  }
}
\newcommand{\roundedboxcyan}[1]{
  \tikz[baseline=(char.base)]{
    \node[anchor=south west, rounded corners, text height=1.5ex, text depth=.25ex, fill=cyan!30, draw=none, text=black, font=\bfseries] (char) {#1};
  }
}
\newcommand{\roundedboxgray}[1]{
  \tikz[baseline=(char.base)]{
    \node[anchor=south west, rounded corners, text height=1.5ex, text depth=.25ex, fill=gray!50, draw=none, text=black, font=\bfseries] (char) {#1};
  }
}

\definecolor{customcolorred}{RGB}{225,159,156}
\definecolor{customcolorgreen}{RGB}{5,204,151}

\newcommand{\boxedred}[1]{
  \tikz[baseline=(char.base)]{
    \node[anchor=south west, rectangle, text height=1.5ex, text depth=.25ex, fill=customcolorred, draw=none, text=black, font=\bfseries] (char) {#1};
  }
}
\newcommand{\boxedgreen}[1]{
  \tikz[baseline=(char.base)]{
    \node[anchor=south west, rectangle, text height=1.5ex, text depth=.25ex, fill=customcolorgreen, draw=none, text=black, font=\bfseries] (char) {#1};
  }
}

\section{Appendix}

\subsection{Overview of the Appendix}
This appendix supplements the proposed \textbf{GeoLens} and our datasets \textbf{GeoMTVR} with details excluded from the main paper due to space constraints.

The appendix is organized as follows:
\begin{itemize}
\item Sec.~\ref{app-details}: More implement details of GeoLens.
    \item Sec.~\ref{app-sec3}: Ablation studies of GeoLens.
    \item Sec.~\ref{app-sec5}: Visualizations of samples and challenging cases.
    \item Sec.~\ref{app-societal-impact}: Potential societal impact of GeoLens and GeoMTVR.
\end{itemize}

\subsection{More Details of Pilot Experiments and GeoLens}
\label{app-details}

Table~\ref{tab:zoomin_qualitative} in Appendix further explains this effect. Solved samples usually depend on evidence from one localized region, such as recognizing a vehicle type, counting objects in a specified port area, identifying a roof color, or determining the relative position of nearby landmarks. In these cases, the main challenge is to locate the relevant region, and one zoom-in operation often provides enough detail. In contrast, persistently failed samples involve global counting, path planning, land-use enumeration, change counting across image pairs, or reasoning over spatially dispersed clues. These tasks require systematic spatial coverage, multi-region comparison, or heterogeneous perceptual capabilities beyond zoom-in. Although the framework allows up to five zoom-in calls and retains the global image in context, limited coverage and the absence of structured exploration still prevent the model from collecting sufficient evidence.

\begin{table*}[h]
\centering
\small
\caption{\textbf{More details of Pilot Experiments.} Qualitative analysis of samples solved by single-tool zoom-in and those that remain persistently failed. Single zoom-in is effective for localized evidence acquisition, but struggles with global coverage, multi-region comparison, and structured spatial reasoning in UHR remote-sensing images.}
\label{tab:zoomin_qualitative}
\setlength{\tabcolsep}{4pt}
\renewcommand{\arraystretch}{1.12}
\begin{tabular}{p{0.16\linewidth} p{0.25\linewidth} p{0.50\linewidth}}
\toprule
\textbf{Group} & \textbf{Representative tasks} & \textbf{Key observation} \\
\midrule
\multicolumn{3}{l}{\textit{Samples solved during RL training}} \\
\midrule
Object classification 
& Vehicle-type recognition 
& A single zoom-in to the target region reveals discriminative visual details, such as the difference between trucks and buses. \\

Regional counting 
& Counting ships in a specified port 
& The relevant region is spatially localized, and objects become clearly separable after zoom-in. \\

Attribute recognition 
& Identifying roof color 
& Zoom-in mitigates the loss of fine-grained appearance caused by global-view downsampling. \\

Local spatial relation 
& Reasoning over nearby landmarks 
& Both targets can be covered by one local crop, making their relative layout directly observable. \\
\midrule
\multicolumn{3}{l}{\textit{Persistently failed samples}} \\
\midrule
Global counting 
& Counting vehicles in a city-scale image 
& Limited zoom-in calls cannot systematically cover the full UHR image, leaving many targets unobserved. \\

Path planning 
& Finding a route between distant points 
& The global view lacks road-level details, while local zoom-in crops are insufficient to form a coherent long-range path. \\

Land-use enumeration 
& Identifying all land-use types 
& The answer depends on multiple heterogeneous regions, but limited local inspections cover only a small subset of them. \\

Change counting 
& Comparing object counts across image pairs 
& Both images require multiple local inspections, making the fixed tool budget insufficient for reliable comparison. \\

Complex reasoning 
& Inferring conditions from dispersed clues 
& The required evidence is distributed across distant regions, and crop-and-zoom lacks a structured exploration mechanism. \\
\bottomrule
\end{tabular}
\vspace{-2mm}
\end{table*}

\subsubsection{Dataset Details of LRS-GRO and XLRS-Bench}

\textbf{LRS-GRO.}
Following LRS-VQA~\citep{lrsvqa}, LRS-GRO~\citep{liu2025zoomearth} is curated from high-resolution remote-sensing imagery sourced from FAIR1M-1.0~\citep{fair1m}, GLH-Bridge, and STAR~\citep{star}. The final benchmark contains 1,224 high-resolution images with an average resolution of approximately 5,000$\times$5,000 pixels, 3,592 bounding boxes, and 13,245 questions. Among them, 1,000 samples are equipped with detailed step-by-step chain-of-thought annotations and are used as the SFT subset. LRS-GRO emphasizes active perception: models are expected to decide whether regional cropping is necessary and to use local visual evidence when the question scope cannot be resolved from the global view.

The benchmark defines 17 major question categories organized into three hierarchical spatial levels. The \textit{Global} level includes counting, season, urban--rural judgment, and scene type. The \textit{Region} level includes counting, existence, status, visual features, function, and category. The \textit{Object} level includes function, material/surface, category, state, relative position, shape/structure, and color/pattern. For region- and object-level tasks, reference bounding boxes are provided to support supervision and evaluation. This hierarchical design enables evaluation of whether a model can adaptively match its visual evidence acquisition strategy to the spatial scope of a question.

\textbf{XLRS-Bench.}
XLRS-Bench~\citep{xlrs-bench} is used as the main ultra-high-resolution remote-sensing VQA benchmark in our experiments. It evaluates both perception and reasoning abilities over extremely large remote-sensing images. As shown in Table~\ref{tab:xlrs-bench-details}, the benchmark is organized into two Level-1 categories, \textit{Perception} and \textit{Reasoning}, and further decomposed into Level-2 and Level-3 sub-tasks. All Level-3 tasks are formulated as multiple-choice VQA questions, covering 3,080 samples in total.

\begin{table*}[htbp]
\renewcommand{\arraystretch}{1.35}
\footnotesize
\caption{\textbf{Characteristics of XLRS-Bench.} The benchmark is used to evaluate perception and reasoning abilities in ultra-high-resolution remote-sensing VQA.}
\centering
\resizebox{0.98\textwidth}{!}{
\begin{tabular}{c|c|ccccc}
\toprule \Gray
\textbf{L1-Task} & \textbf{L2-Task} & \textbf{L3-Task} & \textbf{Abbr.} & \textbf{Annotation Format} & \textbf{Number of Samples} & \textbf{Answer Type} \\ \hline
\multirow{8}{*}{Perception} & \multirow{2}{*}{Counting} & Overall Counting & OC & VQA & 60 & Multiple Choice (A/B/C/D) \\
 & & Regional Counting & RC & VQA & 100 & Multiple Choice (A/B/C/D) \\
\cline{2-7}
 & \multirow{2}{*}{Scene Classification} & Overall Land Use Classification & OLUC & VQA & 100 & Multiple Choice (A/B/C/D) \\
 & & Regional Land Use Classification & RLUC & VQA & 200 & Multiple Choice (A/B/C/D) \\
\cline{2-7}
 & Object Spatial Relationship & Object Spatial Relationship & OSR & VQA & 500 & Multiple Choice (A/B/C/D) \\
\cline{2-7}
 & \multirow{3}{*}{Object Properties} & Object Classification & OCC & VQA & 800 & Multiple Choice (A/B/C/D) \\
 & & Object Color & OCL & VQA & 800 & Multiple Choice (A/B/C/D) \\
 & & Object Motion State & OMS & VQA & 60 & Multiple Choice (A/B for Yes/No) \\
\hline
\multirow{5}{*}{Reasoning} & Route Planning & Route Planning & RP & VQA & 100 & Multiple Choice (A/B/C/D) \\
\cline{2-7}
 & Anomaly Reasoning & Anomaly Detection and Interpretation & AD & VQA & 100 & Multiple Choice (A/B/C/D) \\
\cline{2-7}
 & \multirow{2}{*}{Complex Reasoning} & Environmental Condition Reasoning & ECR & VQA & 100 & Multiple Choice (A/B/C/D) \\
 & & Counting with Complex Reasoning & CCR & VQA & 100 & Multiple Choice (A/B/C/D) \\
\cline{2-7}
 & Spatiotemporal Reasoning & Regional Counting with Change Detection & RCCD & VQA & 60 & Multiple Choice (A/B/C/D) \\
\bottomrule
\end{tabular}
}
\label{tab:xlrs-bench-details}
\vspace{-2mm}
\end{table*}

\subsubsection{More Details of GeoLens}

\textbf{Model architecture.}
GeoLens is based on the Qwen2.5-VL architecture, specifically \texttt{Qwen2\_5\_VLForConditionalGeneration}. The language backbone contains 28 transformer layers with hidden size 3584, intermediate size 18944, 28 attention heads, and 4 key-value heads. The vocabulary size is 152,064. The vision encoder contains 32 layers with hidden size 1280, 16 attention heads, patch size 14, and spatial merge size 2. Visual features are projected into the 3584-dimensional language hidden space. Overall, GeoLens follows the standard Qwen2.5-VL multimodal design, coupling a ViT-style visual encoder with an autoregressive language model.

GeoLens is formulated as an agentic remote-sensing vision-language model. Given an ultra-high-resolution remote-sensing image and a question, the model generates an interleaved trajectory consisting of brief reasoning, structured tool calls, visual observations, and the final answer. It can invoke three tools: \texttt{crop\_and\_zoomin}, \texttt{grounding}, and \texttt{line}. These tools support progressive local inspection, object-level localization, and auxiliary-line-based spatial reasoning. Returned observations are appended to the context so that subsequent reasoning can condition on newly acquired visual evidence.

\textbf{Supervised fine-tuning stage.}
The RTAL stage starts from a supervised checkpoint trained from Qwen2.5-VL-7B-Instruct. The SFT stage uses 13,078 cold-start multimodal instruction samples from three sources. These samples provide multi-tool reasoning trajectories and teach the model the basic format of tool invocation, evidence interpretation, and final answer generation.

\textbf{RTAL stage.}
GeoLens is then optimized with RTAL from the SFT checkpoint. The implementation is based on \texttt{verl}, using 8 GPUs, vLLM rollout, and FSDP actor/reference execution. The RTAL training data contains 9,979 prompts from SuperVQA and LRS-GRO, with an additional validation set of 244 SuperVQA samples. Tables~\ref{tab:geolens-sft-data} and~\ref{tab:geolens-rl-data} summarize the SFT and RTAL data, respectively.

\begin{center}
\begin{minipage}[t]{0.48\textwidth}
\centering
\scriptsize
\captionof{table}{\textbf{SFT data used for GeoLens.} The SFT stage uses 13,078 cold-start multimodal instruction samples.}
\label{tab:geolens-sft-data}
\setlength{\tabcolsep}{4pt}
\renewcommand{\arraystretch}{1.08}
\begin{tabular}{p{0.66\linewidth}|r}
\toprule
\textbf{Source} & \textbf{Samples} \\
\midrule
SuperVQA cleaned/accepted data & 9,211 \\
LRS-GRO zoom-based open-answer data & 3,479 \\
route\_visual\_accepted\_388 & 388 \\
\midrule
Total & 13,078 \\
\bottomrule
\end{tabular}
\end{minipage}
\hfill
\begin{minipage}[t]{0.48\textwidth}
\centering
\scriptsize
\captionof{table}{\textbf{RTAL training data used for GeoLens.}}
\label{tab:geolens-rl-data}
\setlength{\tabcolsep}{4pt}
\renewcommand{\arraystretch}{1.08}
\begin{tabular}{p{0.66\linewidth}|r}
\toprule
\textbf{Source} & \textbf{Samples} \\
\midrule
SuperVQA & 8,066 \\
LRS-GRO & 1,913 \\
\midrule
Total & 9,979 \\
\bottomrule
\end{tabular}
\end{minipage}
\vspace{-1mm}
\end{center}

The SFT training is performed with LLaMA-Factory in a full fine-tuning setting for the language model. The vision tower and multimodal projector are frozen, while the language model is updated. The training uses 8 GPUs with DeepSpeed ZeRO-3, bf16 precision, per-device batch size 1, and gradient accumulation 4, resulting in a global batch size of 32. The learning rate is set to $1\times10^{-5}$ with a cosine scheduler and warmup ratio 0.1. The model is trained for 3 epochs with cutoff length 32768 and maximum image pixels 2,359,296. The final SFT run reaches 1224 optimization steps with a training loss of approximately 0.1199. Tables~\ref{tab:geolens-sft-config} and~\ref{tab:geolens-rl-config} list the main SFT and RTAL hyperparameters.

\begin{center}
\begin{minipage}[t]{0.48\textwidth}
\centering
\scriptsize
\captionof{table}{\textbf{SFT training configuration of GeoLens.}}
\label{tab:geolens-sft-config}
\setlength{\tabcolsep}{3pt}
\renewcommand{\arraystretch}{1.05}
\begin{tabular}{p{0.48\linewidth}|p{0.43\linewidth}}
\toprule
\textbf{Parameter} & \textbf{Value} \\
\midrule
Initialization & Qwen2.5-VL-7B-Instruct \\
Training framework & LLaMA-Factory \\
Trainable modules & Language model \\
Frozen modules & Vision tower and multimodal projector \\
GPUs & 8 \\
Distributed strategy & DeepSpeed ZeRO-3 \\
Precision & bf16 \\
Per-device batch size & 1 \\
Gradient accumulation & 4 \\
Global batch size & 32 \\
Learning rate & $1\times10^{-5}$ \\
Scheduler & Cosine \\
Warmup ratio & 0.1 \\
Epochs & 3 \\
Cutoff length & 32768 \\
Maximum image pixels & 2,359,296 \\
Optimization steps & 1224 \\
Final training loss & $\approx$0.1199 \\
\bottomrule
\end{tabular}
\end{minipage}
\hfill
\begin{minipage}[t]{0.48\textwidth}
\centering
\scriptsize
\captionof{table}{\textbf{RTAL training configuration of GeoLens.}}
\label{tab:geolens-rl-config}
\setlength{\tabcolsep}{3pt}
\renewcommand{\arraystretch}{1.05}
\begin{tabular}{p{0.48\linewidth}|p{0.43\linewidth}}
\toprule
\textbf{Parameter} & \textbf{Value} \\
\midrule
Algorithm & RTAL with GRPO-style policy optimization \\
Implementation & \texttt{verl} \\
Rollout backend & vLLM \\
Actor/reference execution & FSDP \\
GPUs & 8 \\
Rollout number $n$ & 8 \\
Train batch size & 32 \\
PPO mini-batch size & 32 \\
Actor PPO micro-batch per GPU & 1 \\
Max prompt length & 8192 \\
Max response length & 1024 \\
Actor learning rate & $1\times10^{-6}$ \\
KL loss & Disabled \\
Entropy coefficient & 0 \\
vLLM tensor parallel size & 2 \\
vLLM GPU memory utilization & 0.6 \\
Max batched tokens & 16384 \\
Save frequency & Every 10 steps \\
Checkpoint keep limit & 2 \\
\bottomrule
\end{tabular}
\end{minipage}
\vspace{-1mm}
\end{center}

\textbf{RTAL attention regularization.}
During RTAL, the attention regularization is applied to tool-action-related spans and valid generated response tokens through attention JSD terms. The tool-action mask is constructed from spans marked by \texttt{Action} and the corresponding closing delimiter, targeting the region where the model decides which tool to call and how to parameterize it. The response mask covers valid generated tokens. This design concentrates optimization on high-impact reasoning and tool-use regions while preserving the standard policy-optimization objective for answer correctness.

\textbf{Prompting setup.}
The RTAL prompt frames GeoLens as a remote-sensing visual reasoning assistant capable of calling \texttt{crop\_and\_zoomin}, \texttt{grounding}, and \texttt{line}. The system prompt requires brief reasoning before tool calls, normalized coordinates in $[0,1000]$, and final answers inside \texttt{<answer>...</answer>}. The user prompt asks the model to solve the remote-sensing question using visual evidence, optionally invoking tools when necessary, while allowing direct answering when the initial evidence is sufficient.

\subsection{Ablation Studies of GeoLens}
\label{app-sec3}

\begin{table*}[htbp]
\footnotesize
\caption{
\textbf{Data ablation on tool-augmented SFT on XLRS-Bench.} 'Avg.' represents the average accuracy across sub-tasks.
}
\centering
\resizebox{\textwidth}{!}{
\begin{tabular}{l|ccccccccccccc|c}
\toprule
\Gray
\textbf{SFT Data} & \textbf{OC} & \textbf{RC} & \textbf{OLUC} & \textbf{RLUC} & \textbf{OCC} & \textbf{OCL} & \textbf{OMS} & \textbf{OSR} & \textbf{AD} & \textbf{ECR} & \textbf{RP} & \textbf{RCCD} & \textbf{CCR} & \textbf{Avg.} \\
\midrule
w/o Tool Data (SuperRSVQA) & 31.7 & 39.0 & 37.0 & 76.0 & 41.2 & 38.0 & 63.3 & 30.0 & 72.0 & 80.0 & 33.0 & 45.0 & 50.0 & 48.9~\textcolor{red}{$\downarrow$ +0.7} \\
\midrule
SuperRSVQA w/ Zoom-in Tool & 31.7 & 36.0 & 43.0 & 73.5 & 42.5 & 32.0 & 60.0 & 30.0 & 75.0 & 79.0 & 52.0 & 40.0 & 50.0 & 49.6~\textcolor{red}{$\downarrow$ +0.7} \\
\midrule
\textbf{SuperRSVQA w/ Full Tool Data} & 31.7 & 40.0 & 29.0 & 74.5 & 40.1 & 33.1 & 66.7 & 42.0 & 71.0 & 81.0 & 57.0 & 38.3 & 50.0 & 50.3 \\
\bottomrule
\end{tabular}
}
\label{tab:data-ablation-tool-sft}
\end{table*}

\textbf{Data Ablation on Tool-Augmented SFT.}
To systematically analyze the effect of tool-related data during supervised fine-tuning, we construct three progressively enhanced variants based on SuperRSVQA, as shown in Table~\ref{tab:data-ablation-tool-sft}. The first setting uses only the original SuperRSVQA data and is denoted as w/o Tool Data. The second setting further introduces zoom-in local-crop operations into SuperRSVQA, providing single-tool active-perception supervision. The third setting augments SuperRSVQA with the full tool set, including progressive zoom-in, object grounding, and auxiliary-line-based spatial reasoning. Under the same evaluation protocol on XLRS-Bench, performance improves as tool information is gradually introduced. Zoom-in data strengthens local-region perception and benefits fine-grained recognition and regional reasoning, while the complete multi-tool chain further improves multi-step reasoning and complex decision-making. This indicates that tool-augmented data provides not only additional supervision but also structural guidance for learning how to acquire and integrate visual evidence. However, the gain from tool annotations alone remains limited in complex reasoning scenarios, suggesting that an effective learning mechanism is necessary to fully exploit the potential of tool-use data.

\begin{table*}[htbp]
\footnotesize
\caption{
\textbf{Module-level ablation on training stages on XLRS-Bench.} 'Avg.' represents the average accuracy across sub-tasks.
}
\centering
\resizebox{\textwidth}{!}{
\begin{tabular}{l|ccccccccccccc|c}
\toprule
\Gray
\textbf{Training Stage} & \textbf{OC} & \textbf{RC} & \textbf{OLUC} & \textbf{RLUC} & \textbf{OCC} & \textbf{OCL} & \textbf{OMS} & \textbf{OSR} & \textbf{AD} & \textbf{ECR} & \textbf{RP} & \textbf{RCCD} & \textbf{CCR} & \textbf{Avg.} \\
\midrule
Qwen2.5-VL-7B & 33.3 & 37.0 & 26.0 & 71.5 & 30.8 & 37.5 & 66.7 & 33.8 & 65.0 & 69.0 & 25.0 & 38.3 & 50.0 & 44.9~\textcolor{red}{$\downarrow$ +6.8} \\
\midrule
Base Model + SFT & 26.7 & 36.0 & 45.0 & 82.5 & 44.6 & 39.2 & 65.0 & 32.4 & 71.0 & 82.0 & 55.0 & 43.3 & 49.0 & 51.7~\textcolor{red}{$\downarrow$ +2.5} \\
\midrule
\textbf{Base Model + SFT + RTAL (GeoLens)} & 35.0 & 40.0 & 37.0 & 76.5 & 46.5 & 45.6 & 65.0 & 36.8 & 72.0 & 82.0 & 62.0 & 51.7 & 55.0 & 54.2 \\
\bottomrule
\end{tabular}
}
\label{tab:module-ablation-training-stages}
\end{table*}

\textbf{Module-Level Ablation on Training Stages.}
To verify the contribution of each training stage, we compare three settings in Table~\ref{tab:module-ablation-training-stages}: the Qwen2.5-VL-7B base model, the model after supervised fine-tuning, and the final GeoLens model trained with SFT and RTAL. Introducing SFT substantially improves task alignment and basic visual understanding, showing that supervised training provides an effective initialization for remote-sensing tool use. Building on this checkpoint, RTAL further improves performance by optimizing the internal information routing and tool-use decisions during reasoning. This confirms the effectiveness of the progressive training paradigm from base modeling, to supervised alignment, and then to reinforcement-based attention optimization.

\begin{table*}[htbp]
\footnotesize
\caption{
\textbf{Effectiveness of RTAL against token-level RL on XLRS-Bench.} 'Avg.' represents the average accuracy across sub-tasks.
}
\centering
\resizebox{\textwidth}{!}{
\begin{tabular}{l|ccccccccccccc|c}
\toprule
\Gray
\textbf{RL Method} & \textbf{OC} & \textbf{RC} & \textbf{OLUC} & \textbf{RLUC} & \textbf{OCC} & \textbf{OCL} & \textbf{OMS} & \textbf{OSR} & \textbf{AD} & \textbf{ECR} & \textbf{RP} & \textbf{RCCD} & \textbf{CCR} & \textbf{Avg.} \\
\midrule
Base Model + SFT + GRPO & 31.7 & 39.0 & 40.0 & 79.0 & 47.6 & 45.5 & 65.0 & 35.6 & 70.0 & 82.0 & 57.0 & 43.3 & 45.0 & 52.4~\textcolor{red}{$\downarrow$ +1.8} \\
\midrule
\textbf{Base Model + SFT + RTAL (GeoLens)} & 35.0 & 40.0 & 37.0 & 76.5 & 46.5 & 45.6 & 65.0 & 36.8 & 72.0 & 82.0 & 62.0 & 51.7 & 55.0 & 54.2 \\
\bottomrule
\end{tabular}
}
\label{tab:rtal-vs-token-rl}
\end{table*}

\textbf{Effectiveness of RTAL against Token-Level RL.}
To further validate RTAL, we compare it with a standard token-level policy-optimization method, GRPO, under the same training data and initialization conditions. As summarized in Table~\ref{tab:rtal-vs-token-rl}, RTAL consistently outperforms token-level RL in overall performance. Qualitative case analysis further shows that RTAL encourages clearer and more structured reasoning paths during tool invocation, leading to more reasonable tool selection and parameter generation. In contrast, token-level optimization is more prone to redundant calls or unstable decisions. These results suggest that optimizing the model's internal attention distribution provides more effective credit assignment than directly optimizing only the output-token distribution, thereby improving performance on complex reasoning tasks.

\subsection{Visualizations of samples in GeoMTVR}
\label{app-sec5}

\begin{figure}[H]
\centering
\includegraphics[width=0.98\textwidth]{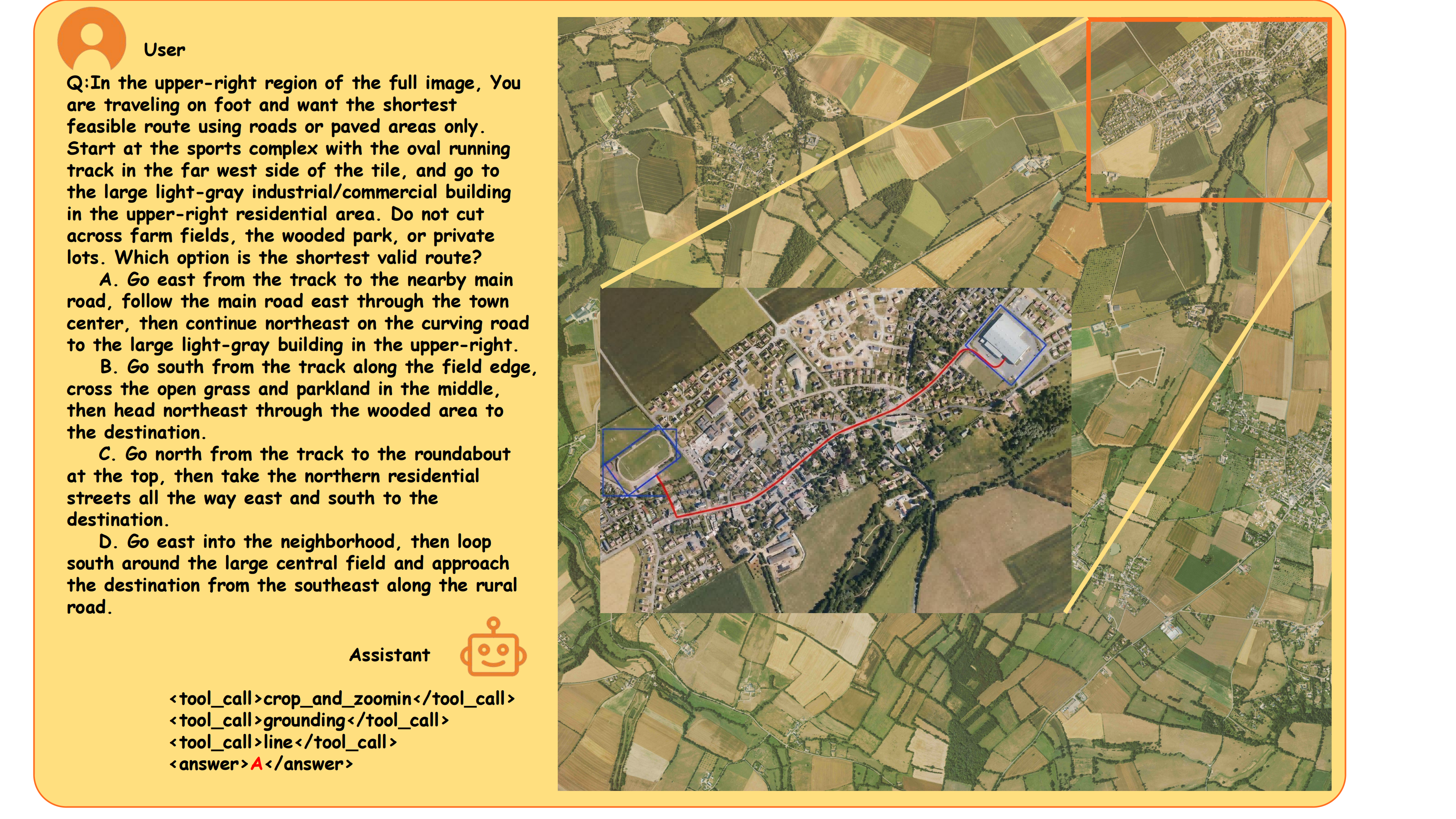}
\caption{\textbf{Visualization of a GeoMTVR sample.} The example illustrates the interleaved reasoning trajectory and visual tool use in GeoMTVR.}
\label{fig:geomtvr-case-01}
\end{figure}

\subsection{Potential Societal Impact}
\label{app-societal-impact}

\textbf{Potential positive societal impact.}
GeoLens and GeoMTVR are designed to improve visual evidence acquisition and reasoning over ultra-high-resolution remote-sensing imagery. Such capabilities can support applications where analysts need to inspect large-area scenes efficiently, such as environmental monitoring, urban planning, land-use analysis, disaster assessment, infrastructure inspection, and ecological observation. Compared with direct global-image reasoning, the multi-tool reasoning paradigm encourages the model to expose intermediate evidence through zoom-in, grounding, and auxiliary-line operations, which may make remote-sensing VQA systems more interpretable and easier to audit. GeoMTVR may also provide a useful research resource for studying active visual reasoning, tool-use behavior, and reliable multimodal decision making under very large image resolutions.

\textbf{Potential negative societal impact.}
The proposed model and dataset are intended for research and decision-support scenarios rather than autonomous high-stakes deployment. As with other remote-sensing analysis systems, incorrect predictions may arise from cloud cover, seasonal changes, outdated imagery, low-quality annotations, or distribution shifts across sensors and geographic regions. Therefore, GeoLens should not be treated as a substitute for expert review in operational settings where errors may have material consequences. In addition, improved analysis of large-area imagery could be misapplied if used without appropriate governance or domain constraints. We therefore recommend that future releases include clear documentation of intended use, evaluation scope, known limitations, and responsible-use terms. Practical deployments should keep humans in the loop, verify important outputs with external evidence, and follow applicable regulations and institutional review procedures.

\end{document}